# Life Style Levels: Neighborhood Delineation using Geospatial Data

Srivatsa Kulkarni[1] and Debarag Banerjee[2]

[1] Senior Data Scientist, L&T Finance
[2] Chief AI & Data Officer, L&T Finance
srivatsakulkarni@ltfs.com

**Abstract.** Fine-scale socioeconomic information is often unavailable across rapidly urbanizing regions of the developing world, like India, limiting the ability to delineate intra-urban variations in affluence and deprivation. This study proposes a scalable, grid-based urban delineation framework using building morphology derived from open-source satellite imagery. Urban areas across 59 Indian cities and towns are partitioned into high-resolution spatial grids and characterized using interpretable morphological indicators, which are combined into a transparent, rule-based scoring framework to delineate areas with contrasting levels of urban affluence. The resulting classifications are validated through ground-level Google Street View observations, revealing a sharp contrast between the grid classes which are consistent with the expected effects of the lifestyle affluence indicators. We further investigate density-based clustering of building footprints in Mumbai to identify dense urban settlements, demonstrating that the resulting clusters exhibit substantial spatial overlap with known informal settlements across the city. Finally, we conduct an exploratory analysis mapping consumer loan delinquency across the derived affluence classes. By relying entirely on publicly available geospatial data, the proposed framework provides a scalable, interpretable, and cost-effective approach for granular urban affluence mapping across Indian cities.

## 1 Introduction

Rapid urbanization has intensified the need for timely and spatially detailed information on the physical and socioeconomic characteristics of cities. Remote sensing has become an indispensable tool for urban studies by providing consistent, repeatable, and synoptic observations of the Earth's surface over large spatial extents (Weng, 2012; Taubenböck et al., 2012). More recently, advances in machine learning, together with the increasing availability of open Earth observation datasets and cloud-based geospatial platforms, have substantially expanded these capabilities by enabling automated extraction of urban features and large-scale spatial analysis (Gorelick et al., 2017;

Reichstein et al., 2019). These developments have supported a wide range of applications, including urban growth monitoring, land-use mapping, poverty estimation, and informal settlement detection (Jean et al., 2016; Kuffer et al., 2016). Despite these advances, fine-scale socioeconomic information remains scarce across much of India, where household income, property valuations, and consumption expenditure are rarely available at neighborhoodscales, while census-based socioeconomic statistics are infrequently updated and reported primarily at administrative units (Census of India, 2011). Consequently, there is growing interest in developing geospatial proxies capable of characterizing neighbourhood-level socioeconomic conditions from remotely sensed observations.

Urban morphology has emerged as an important source of socioeconomic information, with numerous studies demonstrating that characteristics of the built environment are strongly associated with wealth, deprivation, and living conditions. Advances in remote sensing and machine learning have enabled satellite imagery to be used for estimating poverty (Jean et al., 2016), inferring socioeconomic status from urban form (Abitbol & Karsai, 2020), and identifying informal settlements from morphological characteristics (Kuffer et al., 2016; Taubenböck et al., 2018). Across these studies, attributes such as building density, footprint size, settlement compactness, and urban texture consistently capture underlying socioeconomic differences, suggesting that the patterns in the physical structure of cities could be leveraged to delineate neighborhoods.

The Open Buildings dataset developed by Sirko et al. (2021) represents one of the largest publicly available building footprint products, providing extensive coverage across India through building footprints extracted from Sentinel-2 imagery using deep learning techniques. Compared with approaches reliant on expensive commercial satellite imagery, it offers a scalable and cost-effective foundation for urban socioeconomic analysis However, despite the increasing availability of such datasets, limited research has explored whether opensource building footprints can be used for delineating neighbourhood-level affluence across geographically diverse Indian cities. Motivated by this gap, this study addresses the following research question: Can publicly available building footprint data be used to delineate urban neighborhoods into affluent and socioeconomically disadvantaged classes across Indian cities at fine spatial scales?

Capturing neighbourhood-scale socioeconomic variation remains challenging because administrative units rarely correspond to homogeneous urban environments. In Indian cities, a single ward or census unit may simultaneously contain high-rise residential complexes, commercial corridors, planned housing, and densely populated informal settlements, thereby masking substantial intra-urban socioeconomic heterogeneity (Bharathi et al., 2021). Furthermore, standardized neighbourhood boundary datasets are generally unavailable across Indian cities, limiting consistent and reproducible cross-city analyses. These limitations are manifestations of the Modifiable Areal Unit Problem (MAUP), whereby observed spatial patterns and statistical relationships depend on the choice of spatial aggregation units (Openshaw, 1984; Fotheringham & Wong, 1991). To overcome these challenges, we adopt a uniform 100 m × 100 m grid framework that provides a consistent unit of analysis across all cities. By representing every city using the same spatial resolution, the framework enables direct comparison of neighbourhood morphology and affluence distributions while remaining independent of locally defined administrative boundaries.

To address this research question, we propose a fine-resolution grid-based framework for neighborhoods delineation using publicly available building footprints. The framework partitions 59 Indian cities and towns into uniform 100 m × 100 m spatial grids and derives interpretable morphological characteristics—including building count, median building area, and variability in building size—from the Open Buildings dataset. These characteristics are integrated into a transparent, rule-based scoring framework that exploits the inverse relationship between building density and building footprint size to classify the urban landscape into distinct neighborhoods categories. We further investigate a density-based clustering approach to examine whether dense urban settlements can be identified directly from building footprints without predefined labels. Finally, the proposed framework is assessed qualitatively through systematic Google Street View interpretation across representative Indian cities and explored in relation to consumer loan delinquency patterns to demonstrate its potential applications.

The remainder of this paper is organized as follows. Section 2 reviews the literature on remote sensing for urban socioeconomic analysis, building morphology, and spatial urban delineation. Section 3 describes the datasets and study area, including the Open Buildings dataset and customer geocoding pipeline. Section 4 presents the proposed methodology, including the grid construction procedure, rule-based affluence scoring framework, and density-based clustering approach and explores credit default prediction using morphological features. Section 5 presents the empirical evaluation, including the resulting neighborhoods classifications, Google Street View assessment, exploratory analysis of consumer loan delinquency and discusses the limitations of the proposed framework. Section 6 concludes with directions for future research. The Appendix section contains the list of 59 cities and detailed street view snapshots of different cities.

## 2 Related Work

There is an extant literature that leverages the satellite data to draw on ground patterns and explores the socio-economic characteristics of the urban landscape. Maxwell et al. [2] developed a cost-effective approach to derive morphological features from open-source Sentinel-2 imagery, using XGBoost [7] models to identify informal settlements in Africa. Their findings reveal that these settlements possess heterogeneous local characteristics, supporting the hypothesis of varying lifestyle levels, and they developed a generalized model for the aforementioned task which works across multiple cities.

Daniel et al. [8] explored urban delineation by applying DBSCAN clustering to building footprints in Spain, finding that resulting patterns aligned more closely with commute-based trends than with administrative boundaries. Similarly, Ming Liu et al. [9] analyzed factors affecting nighttime lighting patterns, using brightness-based clustering to identify commercial and industrial zones. Their work established significant differences in brightness across various functional zones, including residential, underdeveloped, and industrial areas.

Sungwon Park et al. [10] utilized Nighttime Light (NTL) intensity as a proxy for economic activity to study urban poverty, integrating accessibility, morphological, and economic traits into a multi-dimensional model to estimate poverty indicators. Furthermore, the potential of Large Language Models (LLMs) in identifying lifestyle levels has been explored. Banerjee et al. [11] employed multimodal LLMs to assess living conditions using unstructured data, such as video recordings, achieving an 87% accuracy rate between model-generated ratings and verified human assessments.

Although previous studies have established strong links between urban morphology and socioeconomic conditions, they do not provide a general framework for partitioning entire cities into distinct morphological neighborhood types. We therefore propose a novel grid-based scoring approach that quantifies local building morphology and delineates neighborhoods into dense unplanned settlements, transitional areas, planned residential layouts, and sparsely built or non-residential regions

# 3 Dataset

## 3.1 Open Buildings Dataset

We utilized the third version of the Open Buildings dataset, which contains building footprints developed by W. Sirko et al. (2021). This open-source resource provides estimates for the latitude and longitude coordinates of building polygons and their centroids. Additionally, it includes top-view building area estimates and a confidence score indicating the likely presence of a building. The dataset covers regions in Southern and Eastern Asia, Latin America, and the Caribbean.

These buildings were detected using models such as U-Net [12] trained on 50cm resolution satellite imagery. The objective of these models is not to isolate individual building instances, but to predict a probability score for whether a pixel represents a building, subsequently finding connected components based on a specific probability threshold. These connected components constitute the identified buildings. In below images the building outlines shown in green are the connected components. The Open Buildings data can be utilized to compute regional building density, which is useful for estimating population, environmental impacts such as carbon emissions, and the number of households affected during natural calamities. Attributes of the dataset, with sample values are detailed in Table 1. Figures 1,2,3 below visualize the Open Buildings Dataset.

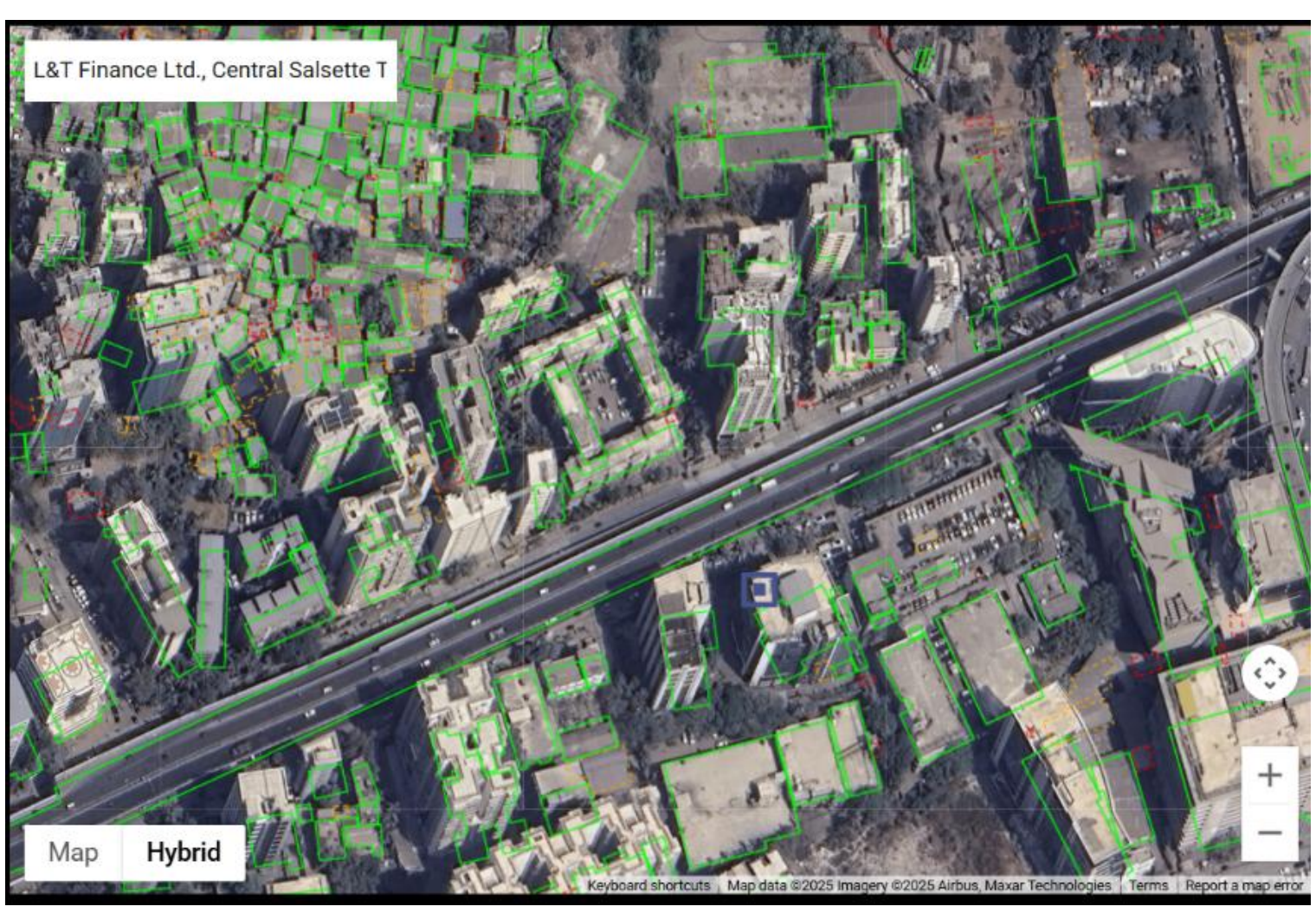

**Fig. 1.** Building Outline Detection

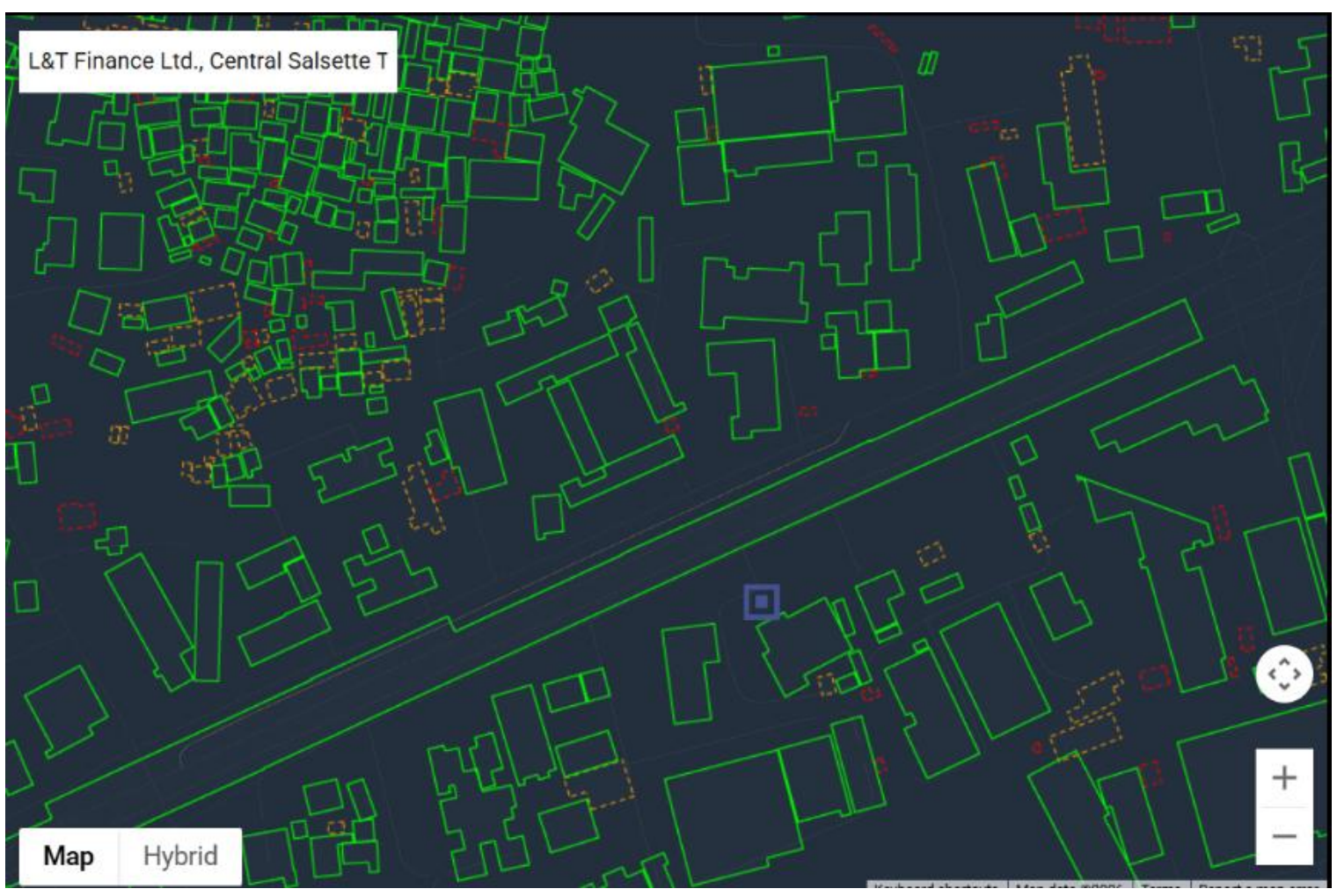

**Fig. 2.** Building projection of the above area

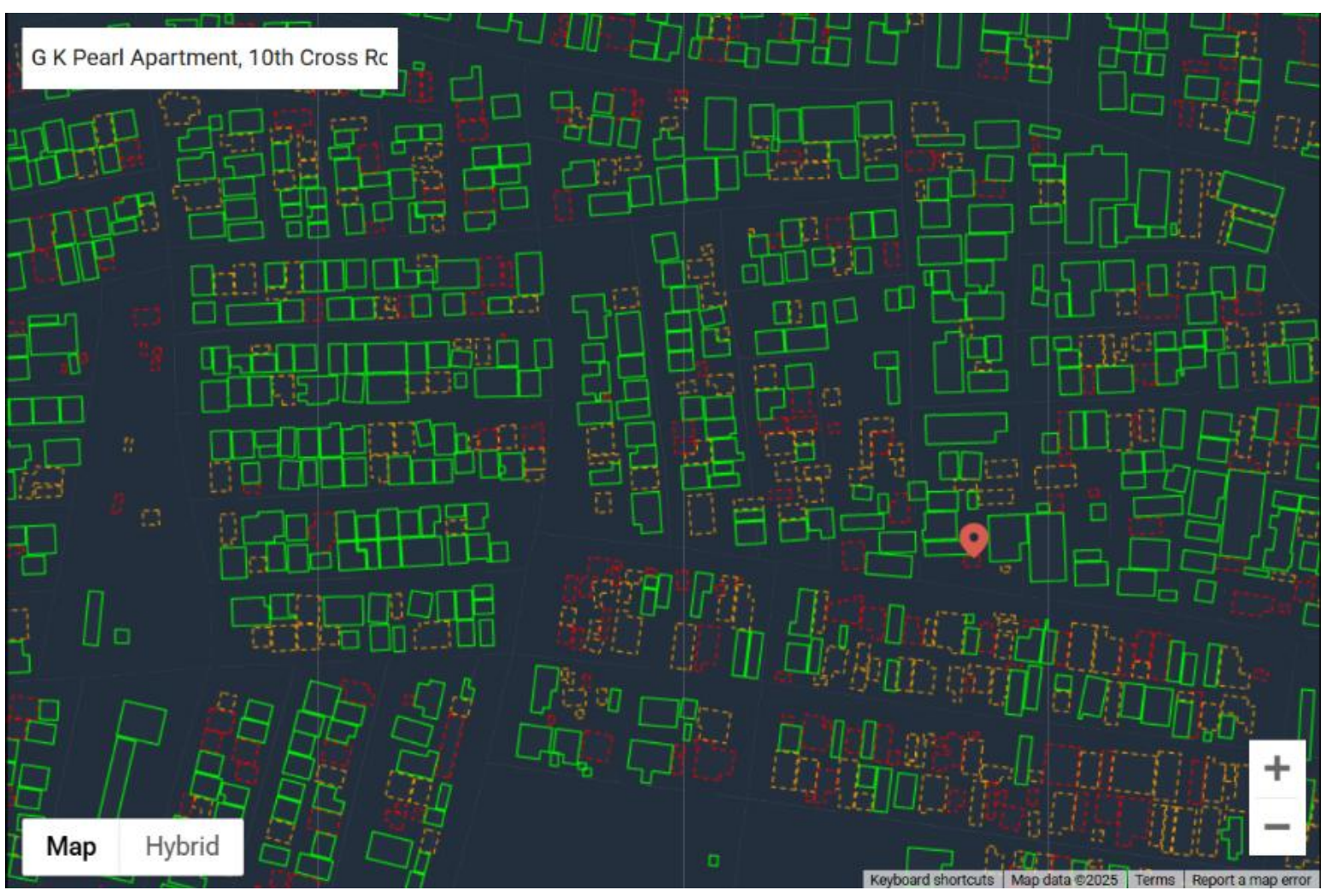

**Fig. 3.** Building Projection: Prime area in Bengaluru which is well planned and spacious

**Table 1.** Open Buildings Dataset

| latitude | longitude | area_in_meters | confi-dence | geometry |
|---|---|---|---|---|
| 19.18669793 | 72.86258626 | 25.5554 | 0.7857 | POLYGON((72.8626235893342 19.186690240989, 72.862615559001 19.1867212066618, 72.8625489314149 19.1867056084846, 72.8625569617593 19.1866746428158, 72.8626235893342 19.186690240989)) |
| 19.2254868 | 72.98028604 | 80.7912 | 0.794 | POLYGON((……)) |
| 19.28752593 | 73.08855179 | 111.1134 | 0.6527 | POLYGON((…….)) |

For any given area of interest, the morphological features of that area will capture the distribution of buildings. Thus, the features would be number of buildings present in that area, the variance of the building area, overall built-up area and also how many small, medium and large sized building footprints are present. These features would uniquely characterize this region. Table 2 contains the details of the derived attributes. The "AREA_BIN_" features were created with an intuitive estimation of average house size in a metropolitan city. The motivation for these features is twofold: first, selected features serve as primary inputs for the grid scoring framework; second, they are utilized as independent variables to evaluate their potential in predicting credit behavior.

**Table 2.** Feature Description

| Column Name | Description |
|---|---|
| ID | Integer Number |
| BUILDING_DENSITY | (Number of building polygons in Region of Interest)/10000 |
| BUILDING_COUNT | Number of building polygons fully contained inside the Region of Interest |
| TOTAL_BUILDING_AREA_(M2) | Total area occupied by the building polygons contained in the grid |
| AVERAGE_BUILDING_SIZE | Mean of all the building area |
| STD_DEV_BUILDING_AREA | Standard deviation of area of all building polygons |
| AREA_BIN_LOW | Proportion of buildings < $40m^2$. Represents small huts, or very dense informal settlements. |
| AREA_BIN_MEDIUM | Proportion of buildings 40 to $190m^2$. Represents standard single-family residential units or small commercial plots. |
| AREA_BIN_HIGH | Proportion of buildings having area >$190m^2$. Represents large apartment blocks, warehouses or civic buildings |
| MEDIAN_BUILDING_SIZE | Median of Building Area. Central tendency measure which is robust to outliers. |

### 3.2 Geographical Scope and Boundary Delineation

The study encompasses 59 Indian cities and towns. (Appendix 10.1). Building footprints were obtained from the Open Buildings dataset for each city using manually curated city boundaries. Initial administrative boundaries from Google Maps were refined through visual inspection to incorporate recent urban expansion and built-up areas omitted from the default boundaries. Figure 4 illustrates this process for Rewari [13], where newly developed peripheral regions excluded from the original administrative boundary were incorporated into the final study extent. The resulting city boundaries were subsequently used to download all building footprints within each urban extent.

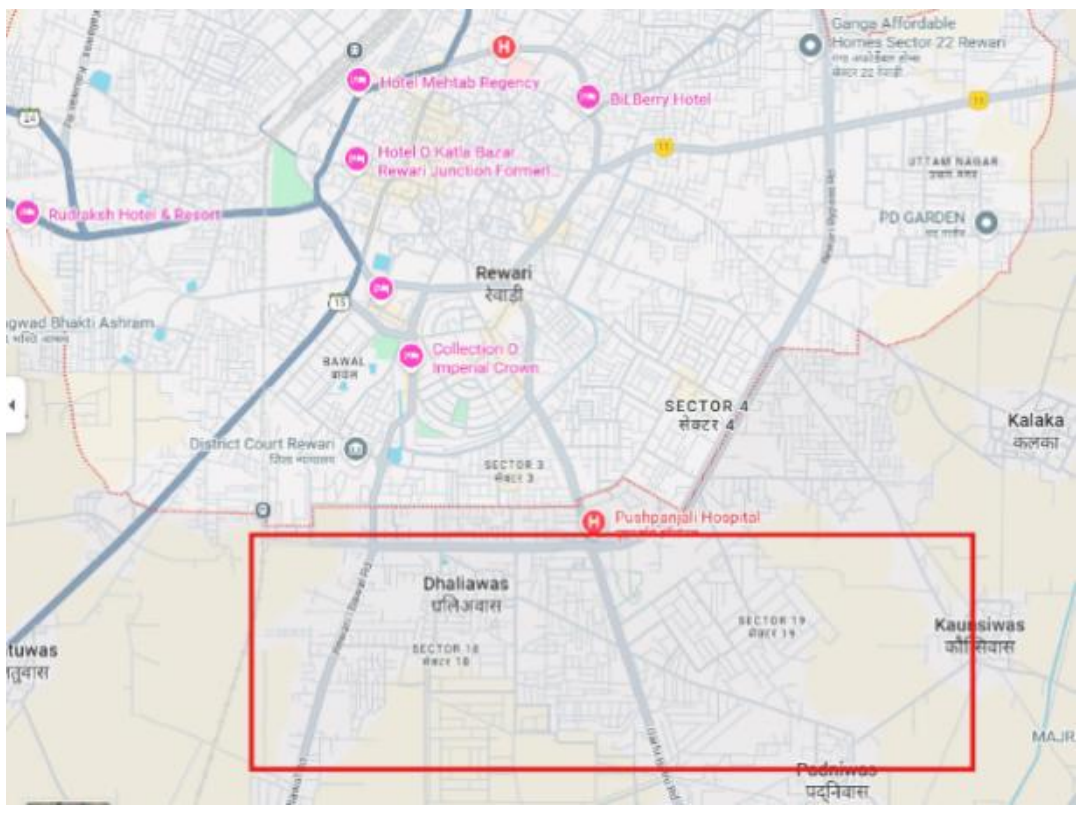

**Fig. 4.** Rewari City Boundary

### 3.3 Socio-Economic Data: Financial Delinquency Markers

To validate the socio-economic implications of urban morphology, this study incorporates anonymized customer loan behavior records sourced from the L&T Finance institution. Residential addresses were converted into spatial coordinates (latitude and longitude) using the Google Maps Geocoding API. These points allow for the mapping of individual financial behavior onto corresponding 100m x 100m spatial grids. In this study, delinquency is defined as a customer being more than 30 days past due (30+ DPD) on paying their loan instalments within the first 12 months of the repayment cycle (12MOB).

# 4 Methodology

## 4.1 Address Scoring

To integrate customer data, residential addresses provided at the time of loan application are converted into geographic coordinates using the Google Geocoding API [14]. The objective is to obtain an approximate coordinate situated within the relevant sub-locality to assign the customer to a specific 100m x 100m grid.

Address components are cleaned of repetitive words and special characters before being processed by the API, which returns candidate addresses broken down into hierarchical components (e.g., zip code, locality, and sub-locality). A match is performed between the API response and the original address using an exact or fuzzy match based on the **Levenstein distance** [15], applying a threshold of 0.7 to signify match strength. To ensure high-definition accuracy, zip codes and sub-localities were assigned higher penalties during the scoring of the geocoded results. The customers are segmented as described in Table 3.

**Table 3.**

| Address score | Description |
|---|---|
| Greater than 50 | These customers will be considered for further analysis. |
| Less than 50 | If the customer address is incomplete or the fuzzy match score with addresses from Google Maps is very low i.e. below 50. |

## 4.2 Spatial Grid Framework

### 4.2.1 Grid Construction

To delineate urban neighborhoods, the study area is partitioned into a systematic framework of small spatial grids. We select a grid resolution of 100m x 100m to strike an optimal balance in capturing the inherent morphological heterogeneity of urban settlements. The objective is to compute the precise spatial coordinates for every constituent grid within the city boundary. Initially, the coordinates of the building polygons are projected into the Universal Transverse Mercator (UTM) coordinate system to establish a 2D geometric framework for accurate distance calculations. We then compute the maximum and minimum X and Y coordinates among all building polygons to derive a rectangular bounding box with vertices at (min(X), min(Y)), (min(X), max(Y)), (max(X), max(Y)), (max(X), min(Y)). Starting from a primary vertex, new grid coordinates are stored by incrementing the X and Y axes by a constant distance of 100m, iterating until the opposite vertex of the bounding box is reached. This approach ensures that we do not overlook local variations in building density, as larger grids might result in the mixing of distinct neighborhood classes.

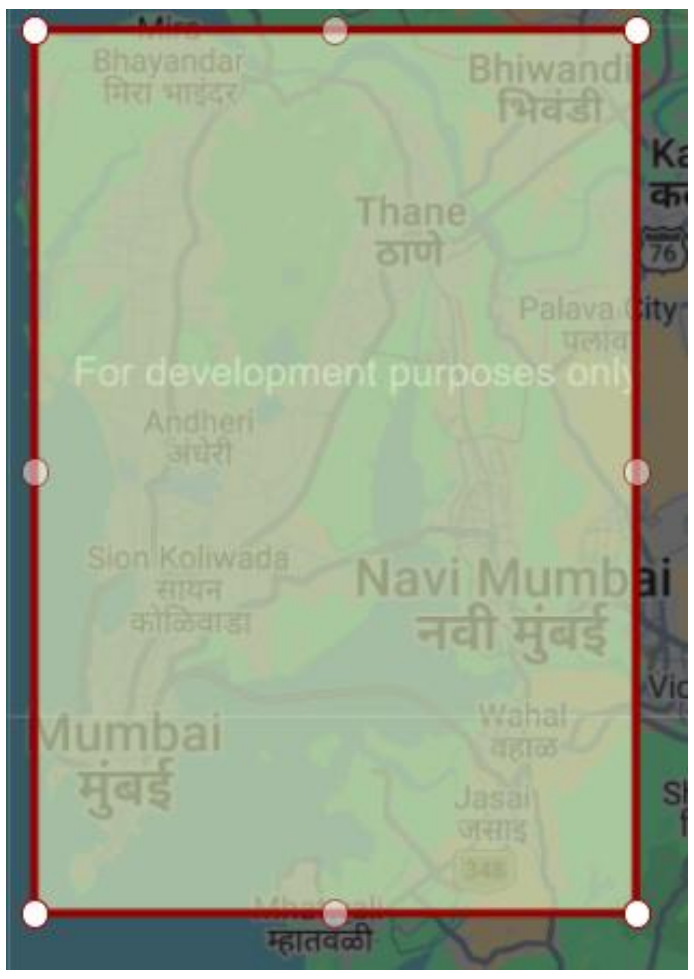

**Fig. 5.** Bounding box for Mumbai.

### 4.2.2 Grid Scoring

Each grid is scored utilizing the building footprint information contained within its boundaries. We first index all building polygons using a Sort-Tile-Recursive (STR) [16] tree. A query is then performed on the tree by passing the grid coordinates as the region of interest, which returns all building polygons within that grid. From this, we derive the total building count and the median building area of all included buildings. In this framework, planned spacious areas are conventionally assigned higher scores relative to high-density regions. This categorization is derived from a critical observation of the variance in building count and building area across distinct neighborhood types. Specifically, planned areas are characterized by a lower relative building count and have footprints of larger size and informal settlements have higher building count and smaller building footprint. To represent the footprint size in each grid, the Median Building Area is utilized as the measure of central tendency. This metric is selected for its robustness against outliers where if a grid contains mix of small and large building footprints median would remain unaffected. Figures 6, 7, 8 show heatmap based on **mean** and **median**. The grid score is formulated based on the direct proportionality of the Median Area and the inverse proportionality of the Building Count, as expressed in Equation (1).

$$Grid\ Score = \frac{Median(Building\ Area)}{Building\ Count\ in\ the\ Grid} \tag{1}$$

For grids containing fewer than two buildings, no score is computed, and they are assigned a distinct "Purple" classification to denote non-residential or sparsely built areas, such as infrastructure and parks. For the remaining urban fabric, we generate a heatmap based on a linear color scale where low scores are represented in red, transitioning in intensity toward a green threshold for higher scores. In Mumbai, scores exhibited a broad range from 844 to 6,201,520. Our empirical evaluation revealed that grids within specific score ranges tended to cluster spatially; notably, the transitions between color intensities occurred at consistent value thresholds. Consequently, we established an upper limit to categorize all grids below a specific value as "Red," effectively isolating the highest-density regions. Table 4 below provides description of the different classes.

A critical observation during the refinement of our scoring model was that the delineation of dense regions appeared significantly sharper when utilizing the Median rather than the Mean in the numerator. This is documented in Figures 6, 7 and 8. To standardize these thresholds, we conducted a manual inspection of a large random sample of grids across nine representative cities to define the thresholds for the Red, Amber, and Green classes. Interestingly, our analysis revealed that Red grids consistently accounted for no more than 5% of the total urban area, while Amber grids never exceeded 9%.

**Table 4.** Grid Label Definition

| Label | Description |
|---|---|
| Red | Represents areas with significantly higher building density relative to the rest of the city. These regions overlap with informal settlements for Mumbai. |
| Amber | These grids often cover, developing residential zones, or edges of parks and infrastructure. |
| Green | These grids cover spacious residential areas with wide roads and planned layouts. |
| Purple | Grids with two or fewer buildings. These usually represent lakes, forests, industrial zones, or major highway intersections. |

Leveraging this statistical consistency, we applied a percentile-based classification for the remaining 50 cities: grids within the 5th percentile are labelled Red, those up to the 9th percentile are labeled Amber, and all subsequent grids are labelled Green. This methodology was validated across 14 cities using Google Street View, where we correlated red grids with narrower streets and high-density small housing, contrasting them with the wider roads and spacious layouts of green grids. We note that this study does not currently implement corrections for low sample sizes in grids that partially overlap with non-residential features like roads or parks.

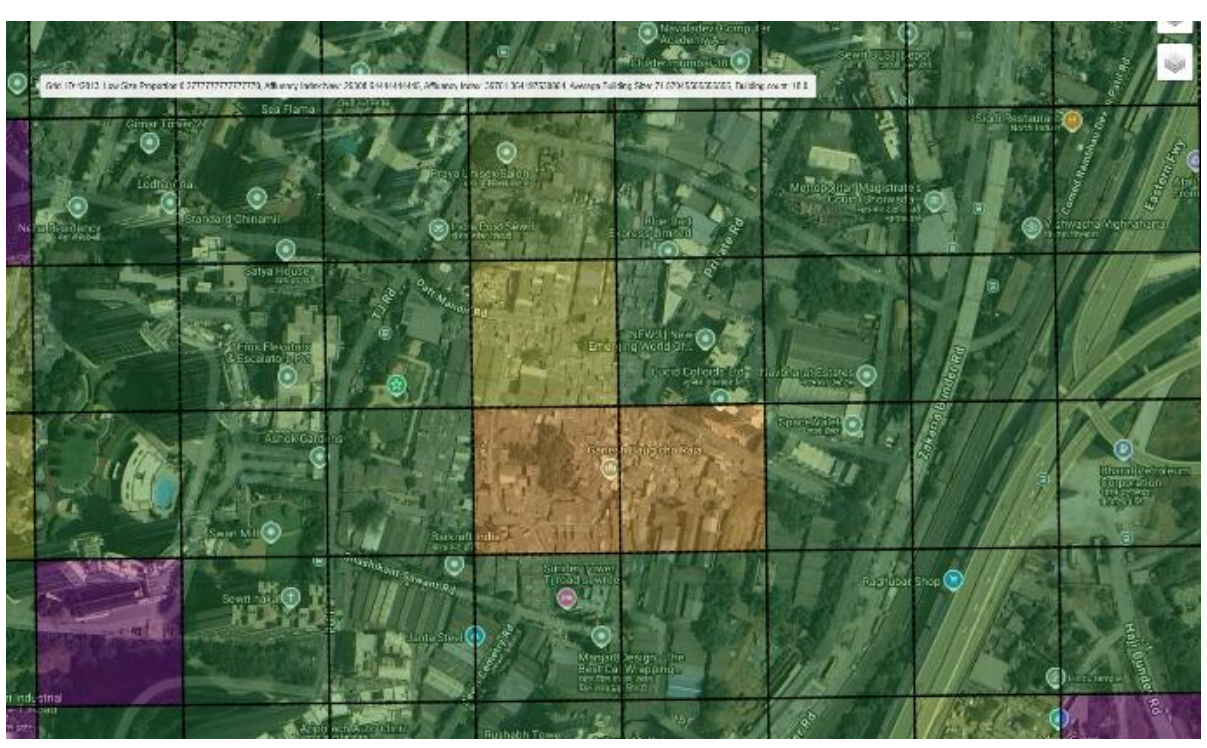

**Fig. 6.** (a)

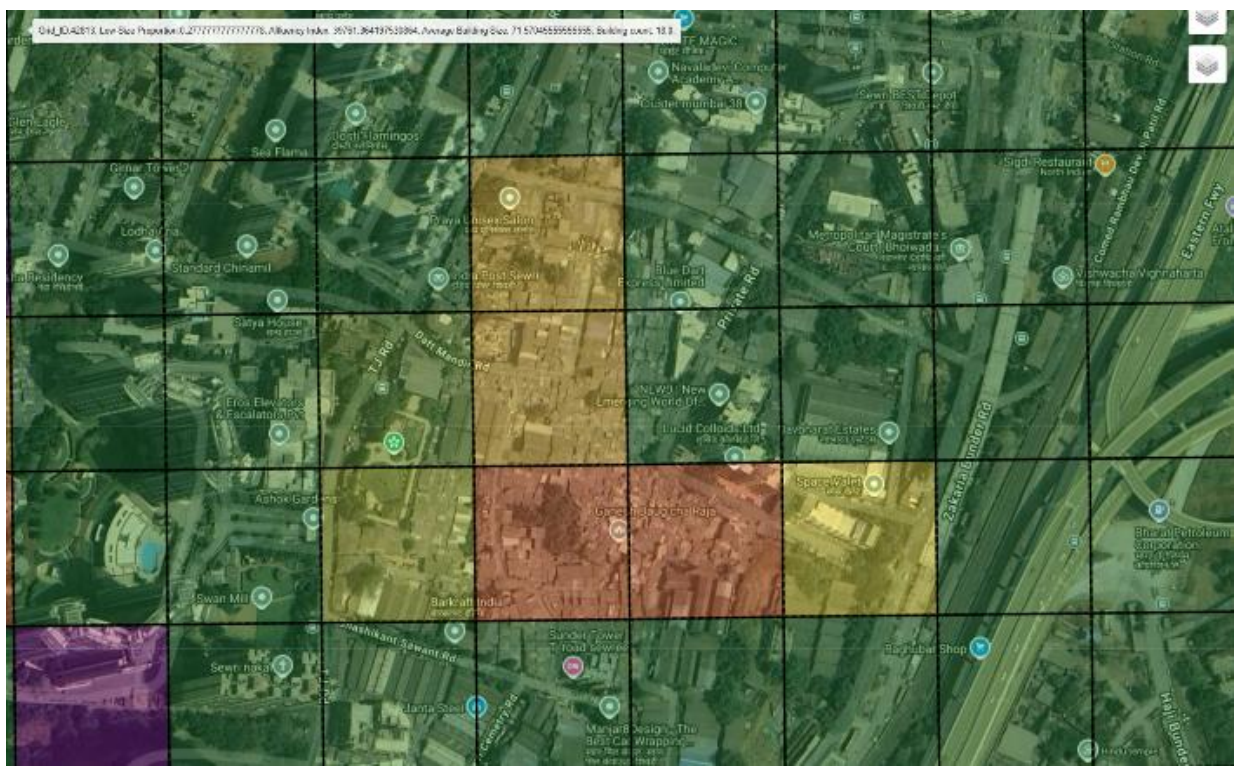

**Fig. 7.** (a)

Figure 6(a) is heatmap based on Mean and Figure 7(a) is based on the Median. Notice that the red color intensity is higher for the Median better capturing the density. Note that the green grids cover the high-rise areas.

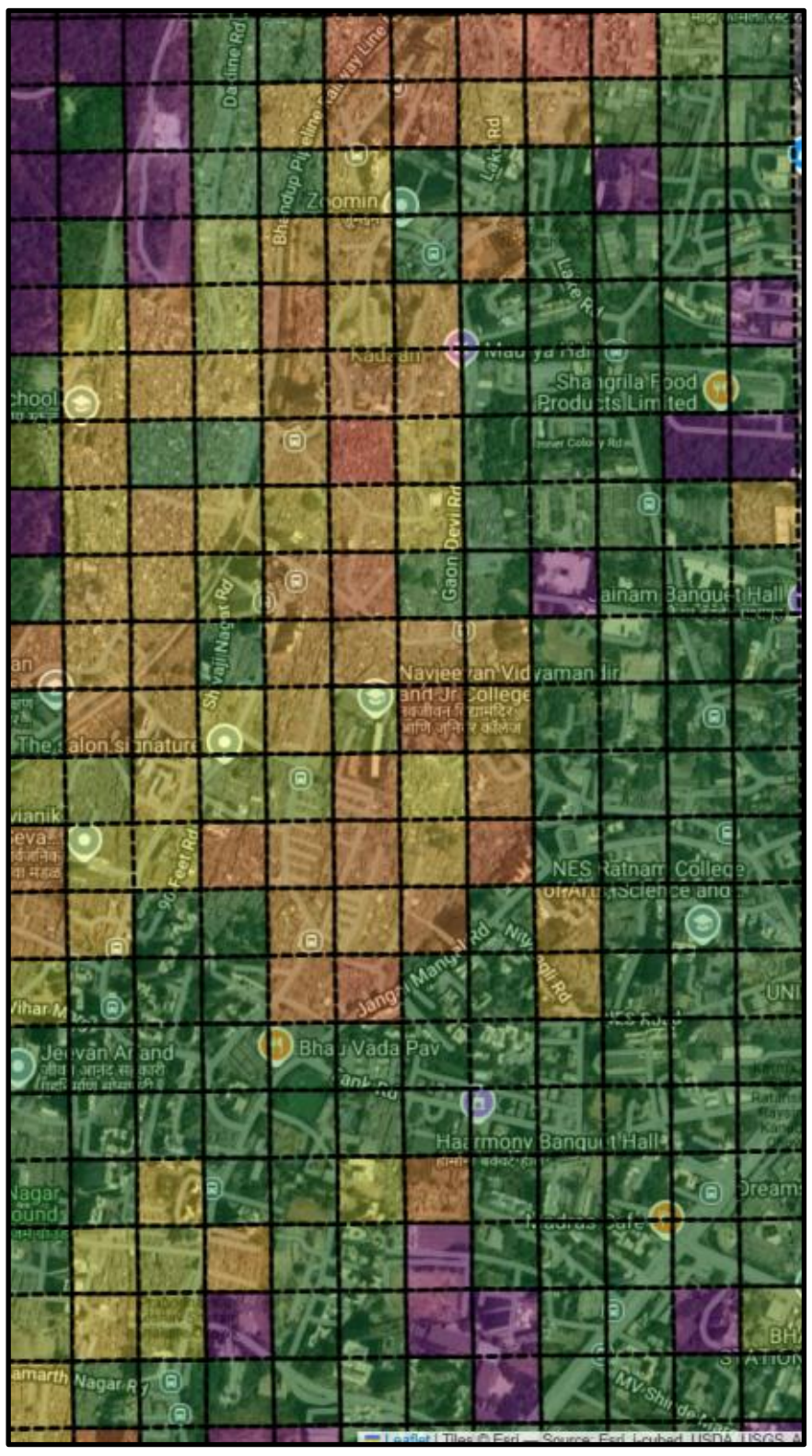

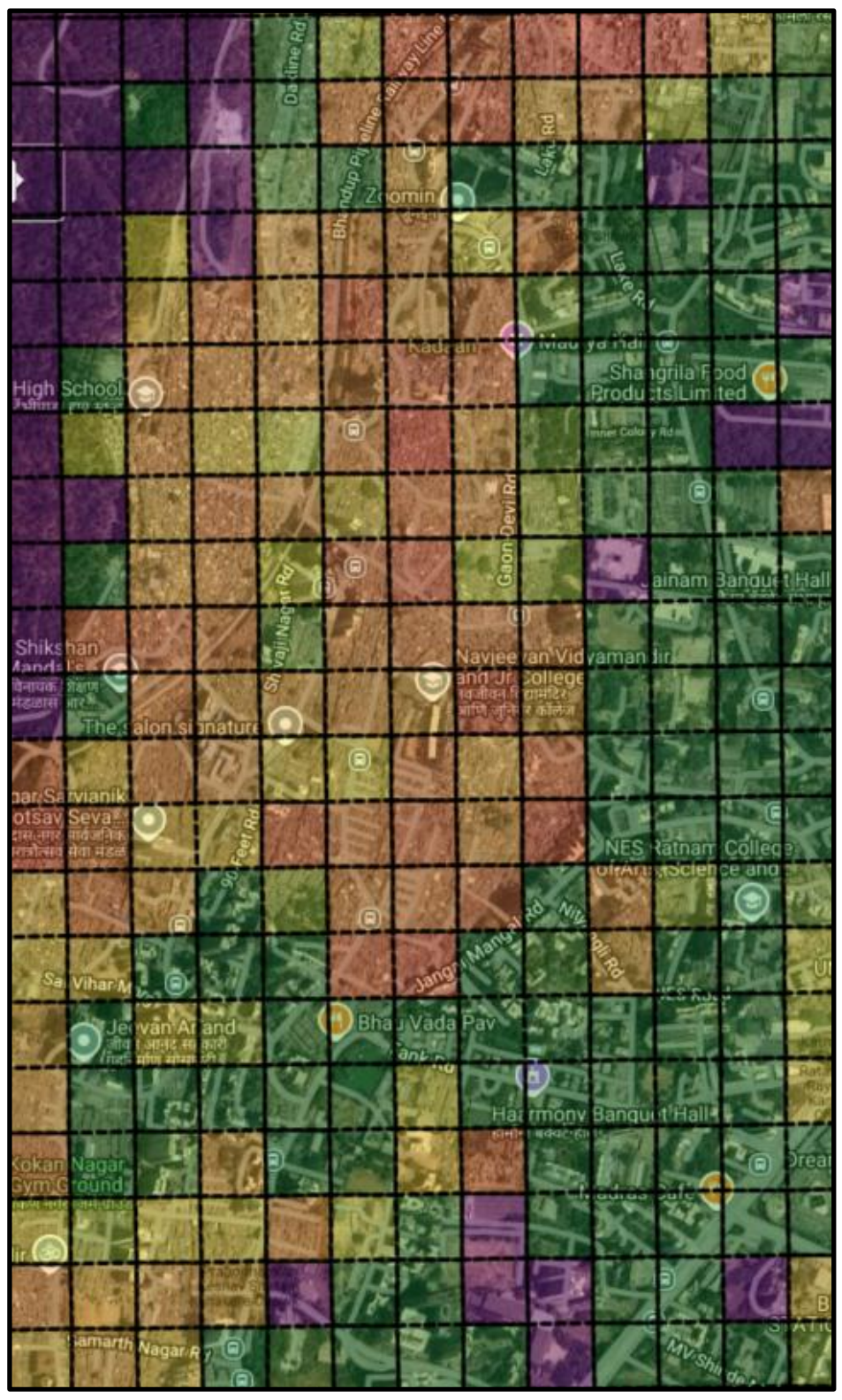

**Fig. 8.** Left image is Mean based and Right image is Median. Notice that the red color intensity is higher for the Median better capturing the dense neighborhoods. The Green areas cover planned and spacious layouts. The Purple grids are having too few buildings

## 4.3 Density Based Neighborhood Classification

Urban environments are frequently characterized by acute space constraints, often giving rise to "less-developed" sectors where housing is aggregated in dense, unplanned configurations. These regions exhibit a structural density that contrasts sharply with the expansive, geometric layouts of planned neighborhoods. Given the inherently spatial dimension of our dataset, we investigate the feasibility of isolating these high-density neighborhoods from the broader urban fabric. The technical challenge requires an analytical framework capable of processing a vast volume of building polygons without necessitating prior descriptive labels or localized neighborhood profiles. This requirement aligns with the capabilities of the DBSCAN (Density-Based Spatial Clustering of Applications with Noise) [1717] algorithm. DBSCAN is uniquely suited for this task as it identifies distinct data classes through specified proximity parameters, requiring minimal domain-specific knowledge of a neighborhood's internal characteristics. Furthermore, the algorithm autonomously determines cluster dimensions and boundaries based on local density variations.

In this context, the objective is to statistically separate coherent clusters from "noise." A cluster is defined as a contiguous collection of data points possessing a higher internal density relative to its exterior surroundings. Conversely, noise points are identified as spatial outliers that exhibit a density lower than the minimum threshold established for cluster formation.

Building polygon centroids from the Open Buildings dataset are utilized as input for the DBSCAN algorithm, where the primary challenge lies in determining the optimal Eps and MinPts parameter values that generalize effectively across the entire city. In this context, MinPts represents the count of building centroids—effectively the number of houses—located within the Eps distance, measured in meters. We employ the Haversine distance metric [18] to account for the Earth's curvature. Following iterative experimentation, manual cluster inspection, and parameter fine-tuning, we established values of Eps = 11 and MinPts = 5 for Mumbai. This logic dictates that if five houses are identified within an 11-meter radius, a seed cluster is formed and subsequently expanded according to DBSCAN principles. To define the spatial extent of these clusters, a unary union operation is performed on the building geometries within each cluster to create a single unified geometry. We then apply a concave hull algorithm [20] to delineate the smallest polygon that encompasses all constituent building geometries. The various stages of this filtering process are summarized in Table 5. A subsequent challenge involves filtering these dense clusters, which is addressed by adapting the previously discussed grid scoring formula. Rather than applying the formula to a fixed grid, we utilize the building count and the Median Building Area specific to each cluster to compute a cluster-level score. A critical observation is that most dense clusters exhibit significantly lower scores compared to smaller clusters associated with planned layouts; this distinction is essential for retaining dense clusters through the application of a score-based threshold.

Clusters are visualized using a heatmap, where low-scoring dense clusters are represented in red and less dense areas in green. A final threshold was established after analyzing the scores of these dense clusters and most of the clusters belonging to planned layouts were filtered. As shown in Figure 10, 12 low-density areas result in smaller clusters, while many clusters in Mumbai demonstrate a striking overlap with informal settlements, Figure 12 to 16. This was verified via satellite imagery by the presence of thatched or tin roofs and closely packed houses. This study was mainly done for Mumbai city. The nature of the areas such as informal settlements or planned layouts, high rise buildings was understood by taking virtual tour of the cluster region using Google Street View.

**Table 5.** Stages of Clustering

| Stage | Number of Data Points |
|---|---|
| Initial Building Geometries covering a large area of Mumbai | 1323307 |
| DBSCAN Clusters | 40,073 |
| Clusters with score <=8000 | 1099 |

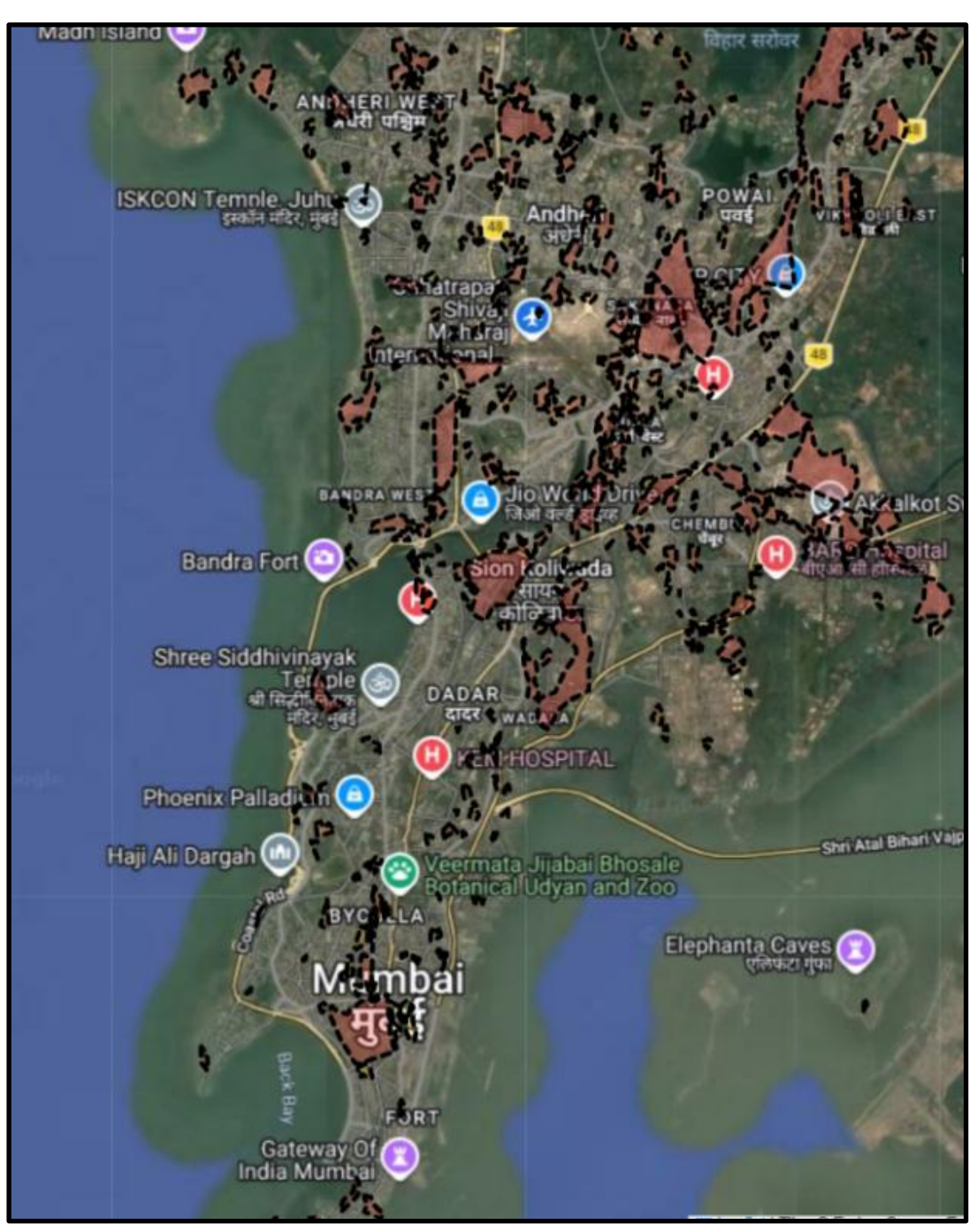

**Fig. 9.** Clustering Results with heatmap based on the cluster score.

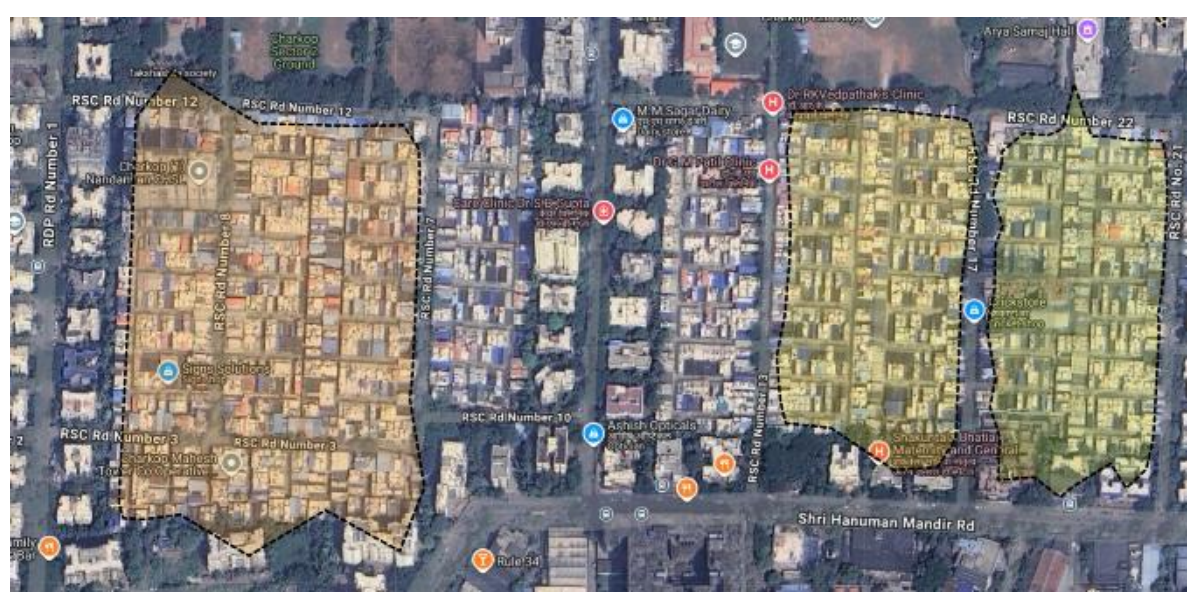

**Fig. 10.** Planned layout, RSC Road, Mumbai

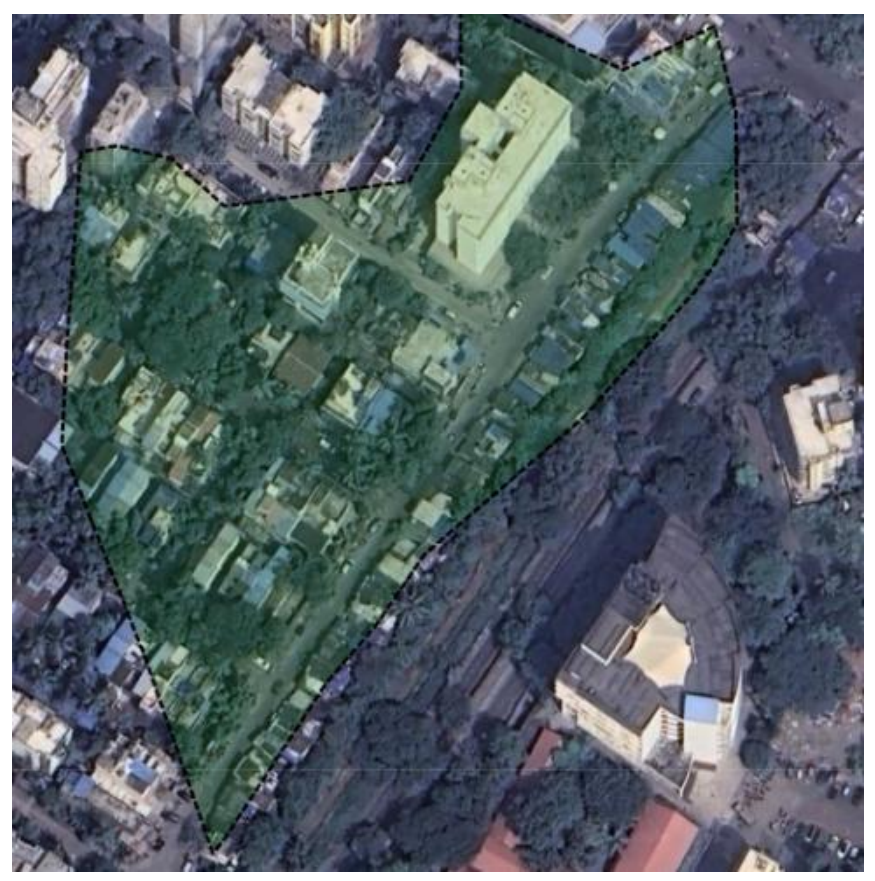

**Fig. 11.** Planned layout

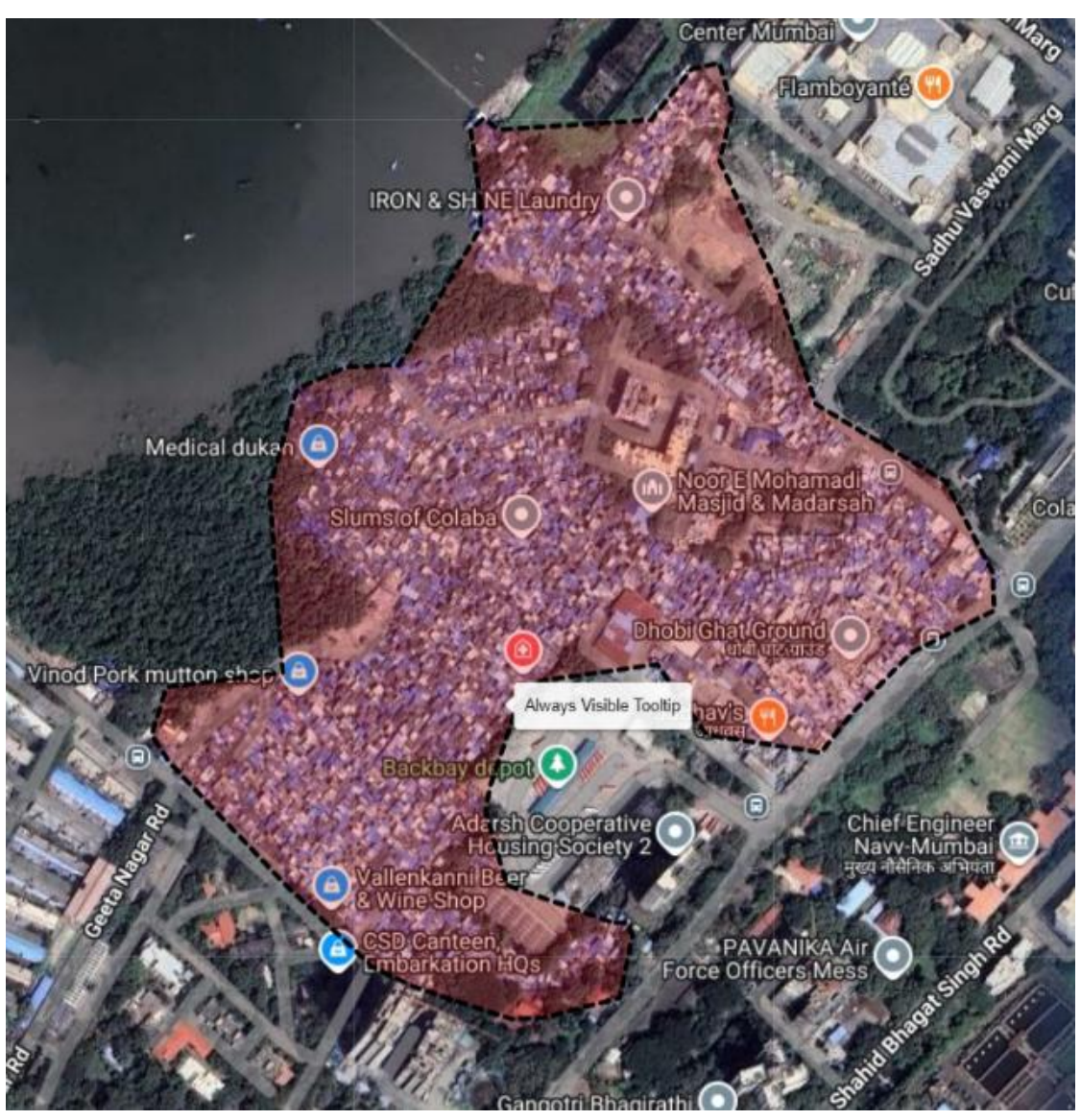

**Fig. 12.** Slums of Colaba

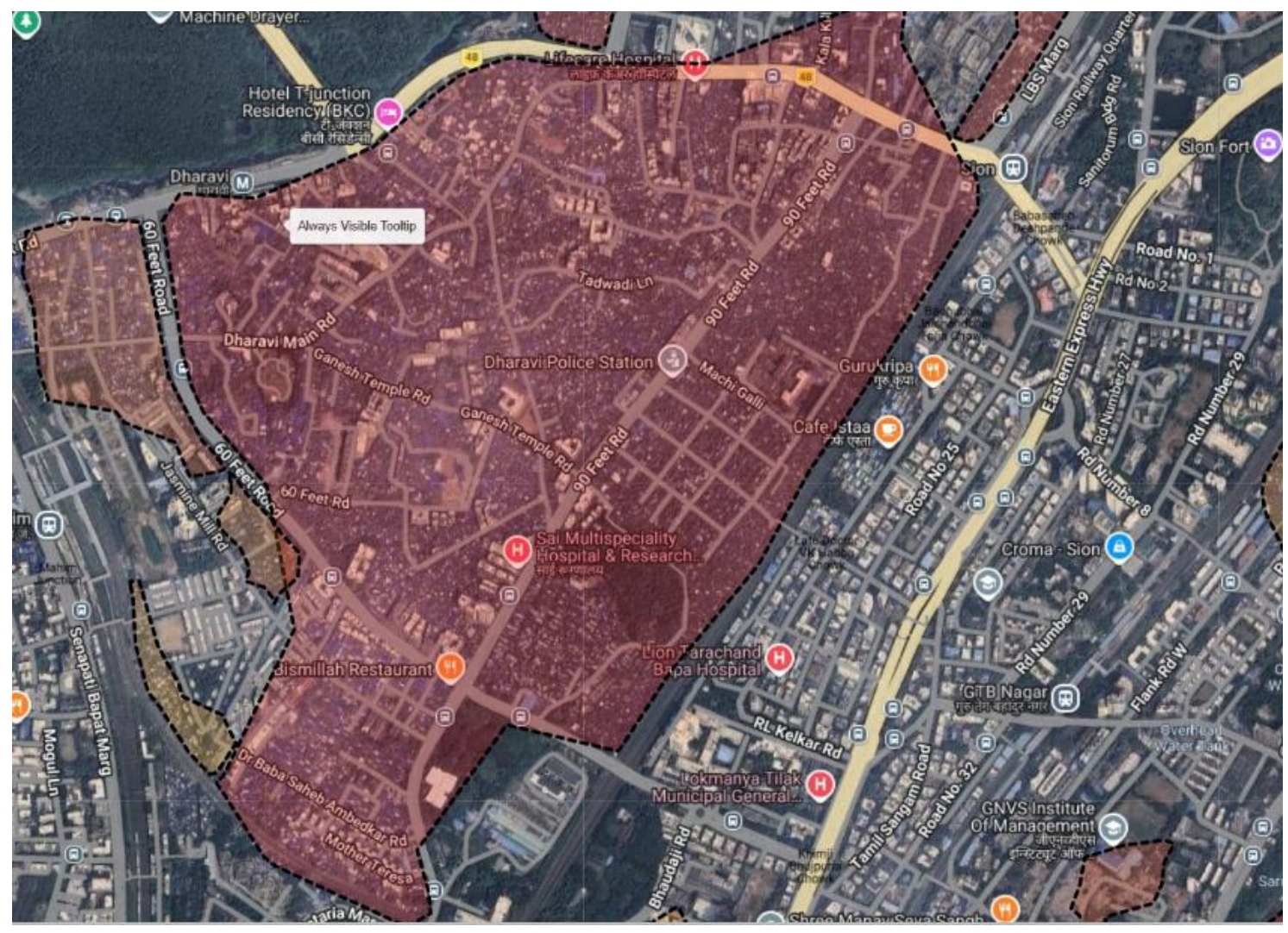

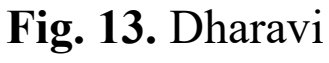

**Fig. 13.** Dharavi

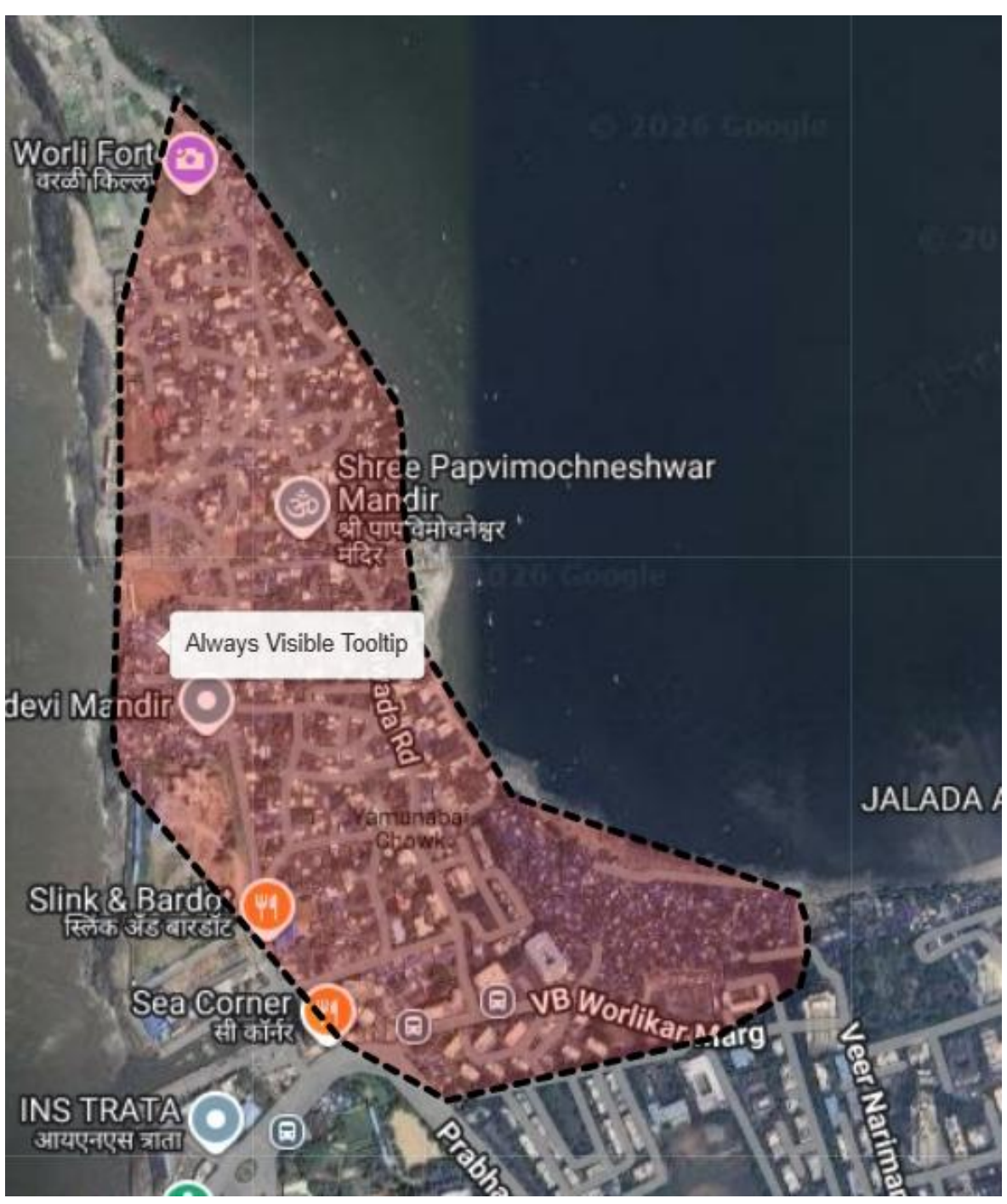

**Fig. 14.** Worli, Mumbai

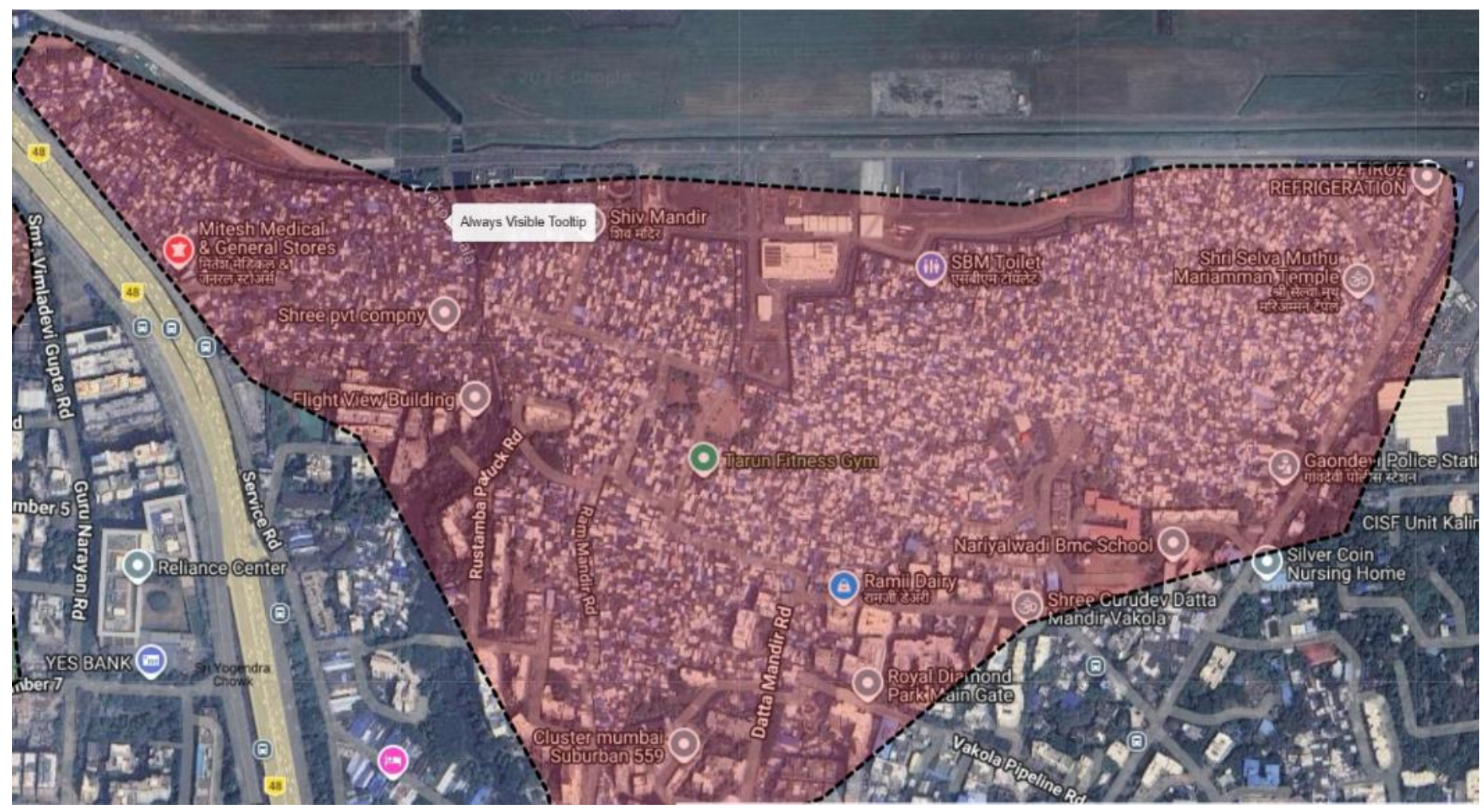

INTERNAL

**Fig. 15.** Near Mumbai Airport

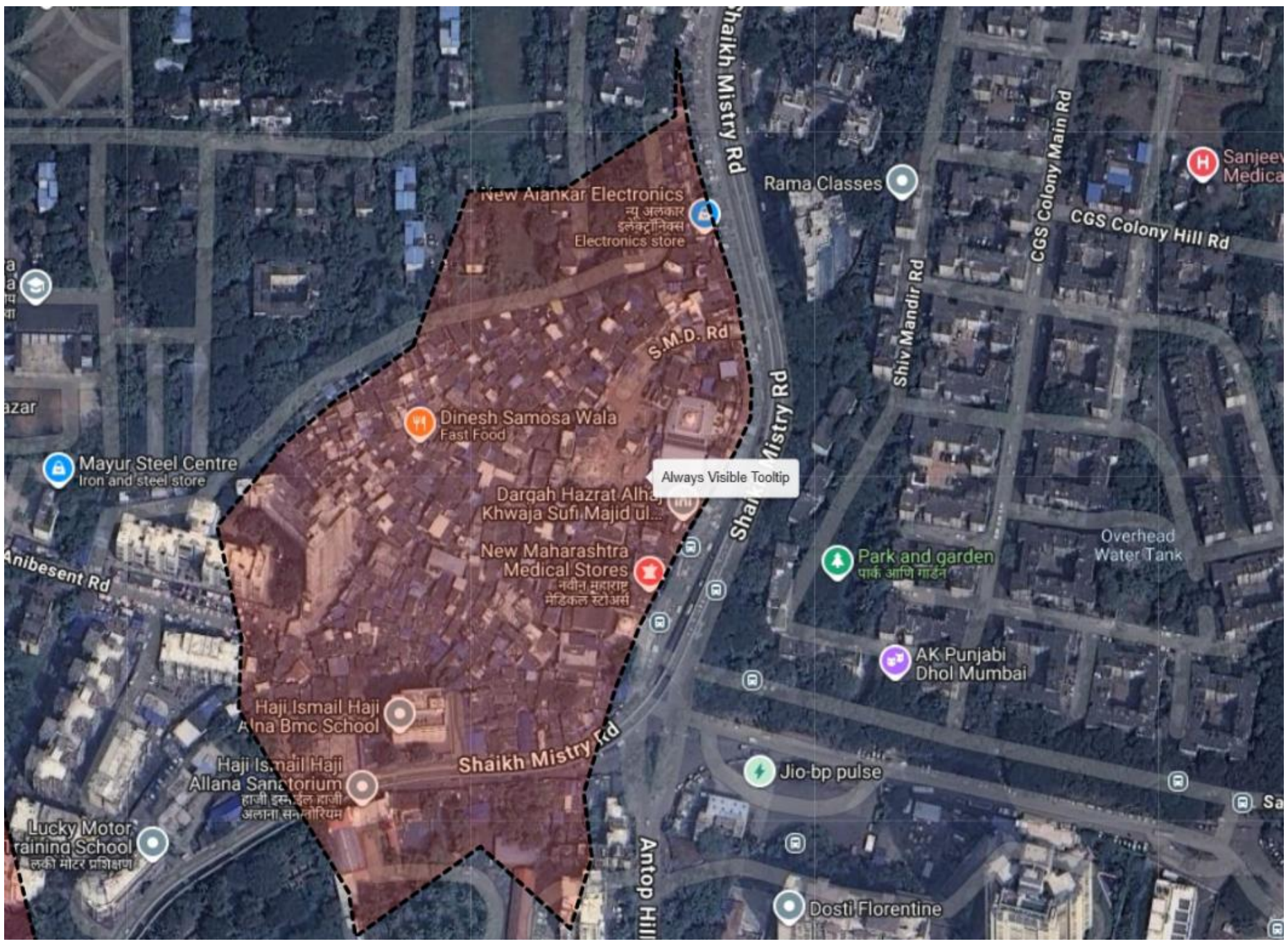

**Fig. 16.** Informal Settlement Near Shaikh Mistry Road, Mumbai

## 4.4 Delinquency Prediction

The primary objective of this phase is to evaluate the effectiveness of the derived morphological features in predicting financial delinquency for customers at the 100m x 100m grid resolution. For each spatial grid, we have extracted several structural features, including building count, median building area, and statistical measures such as the mean and standard deviation of building areas described in Table 2, excluding “ID”, “Building Density” as the latter is directly correlated with “Building Count”.

Furthermore, we incorporated three specific features based on building area thresholds:

- Low Sized Buildings: Structures with an area less than $50m^2$.
- Medium Sized Buildings: Structures with an area greater than $50m^2$ less than $120m^2$

- Large Sized Buildings: Structures with an area greater than $120m^2$ less than $200m^2$

These features represent the proportion of buildings within those specific ranges contained within a given grid. From the available demographic details, the "CITY" attribute was retained and one-hot encoded to account for unique density distributions and varying delinquency behaviors across different urban centers.

To assess the predictive capacity of morphological grid features, the study defines a binary classification task where the objective is to estimate the probability of financial default within a specific timeframe based on a customer's localized urban environment. Default is defined as a customer being more than 30 days past due (30+ DPD) on their loan instalments at any point during the first 12 months of the repayment cycle (12MOB). For instance, this could involve missing the first two consecutive instalments or missing the 7th and 9th months while paying the 8th.

Problem Formulation:

The target variable Y is defined based on a 30+ days past due (DPD) criterion within the first 12 months on book (12MOB). The response for the $i^{th}$customer is formulated as:

$$Y_i = \begin{cases} 1\ if\ customer\ defaults\ (30 + DPD\ within\ 12MOB) \\ 0\ otherwise \end{cases} \quad (2)$$

We performed an Information Value (IV) [209] test to measure the predictive power of these features and found they exhibited a weak influence on the target, with values less than 0.02. Additionally, a Variance Inflation Factor (VIF) analysis confirmed that no multi-collinear features were present, ensuring that redundant information was minimized.

The baseline model is logistic regression [20] used to model the posterior probability of default given a vector of grid-level morphological features **X** (such as building count and median area). The model is expressed through the logit link function:

$$\ln\left(\frac{P(Y = 1|X)}{1 - P(Y = 1|X)}\right) = \beta_0 + \sum_{j=1}^{n} \beta_j X_j \quad (3)$$

Where:

- P(Y=1|X) is the predicted probability of default
- $\beta_0$ is the intercept
- $\beta_j$ *are the structural coefficients*

Model assumptions for Logistic Regression:

- Linearity of the morphological features and logit of outcome

- Independence of Observations: Credit behavior of a person in a grid is not influenced by a neighboring grid.
- The target Y is binary in nature
- Absence of multicollinearity

The model performance is mentioned in the below Table 6. The logit model resulted in a Gini score [21] which is (2*AUC -1) of 14% and a Kolmogorov-Smirnov (KS) [22] statistic of 10.5%. We subsequently explored the XGBoost algorithm, a highly scalable non-parametric model based on tree boosting. To optimize the model, we utilized the Optuna [23] framework for hyperparameter fine-tuning, which resulted in an improved Gini score of 15.06% and a KS value of 10.56%.

**Table 6.** Model Performance

| Model | Gini Score | Kolmogorov-Smirnov (KS) |
| --- | --- | --- |
| Logistic Regression | 14% | 10.5% |
| XGBoost | 15.06% | 10.56% |

# 5 Results and Validation

## 5.1 Grid Results

We have computed the delinquency rates for the Red, Amber, and Green neighborhoods across 59 cities. The description of these classes is provided in Table 4. In this context, delinquency is defined as a customer missing two consecutive EMI instalments within the first 12 months following loan disbursal (30+ 12MOB). Table 7 summarizes these results.
**Monotonic trend** is observed when D(Green) < D(Amber) < D(Red).
**"Close Trend"** is identified when

- D(Green)<D(Amber)=D(Red)
- D(Green)=D(Amber)<D(Red)

Here D(Green) is the delinquency computed as follows:

$$D(Green) = \frac{Number\ of\ Delinquent\ Customers\ from\ Green\ Grid}{Total\ Customers\ of\ Green\ Grid} \quad (4)$$

Similarly, the delinquency is computed for the remaining classes. Table 7 shows the delinquency pattern observed for different business verticals.

**Table 7.** Trend across TW and PL products

| Product | Monotonic Trend | Close Trend | Total |
|---|---|---|---|
| Two-Wheeler Loan (TW) | 17 | 9 | 26 |
| Personal Loan (PL) | 10 | 3 | 13 |
| TW+PL | 13 | 10 | 23 |

For several cities, the results indicate a clear ordinal relationship with the lifestyle index (Red > Amber > Green), while irregular ordering was observed in others. The Purple class, which covers parks, roads, and areas with negligible building counts, was excluded as it likely does not represent a specific lifestyle level. Table 8 displays the delinquency details for Red, Amber, and Green neighborhoods for selected cities, where Grid Delinquency is expressed as a factor X of the overall City Delinquency.

$$Grid\ Delinquency = (City_Delinquency\ * X)$$

where X is the value in Table 8.

**Table 8.** City wise Delinquency Ratio[1]

| STATE_CITY | Red Grid | Amber Grid | Green Grid |
|---|---|---|---|
| **KARNATAKA_BENGALURU** | **1.249** | **1.068** | **0.966** |
| DELHI_NEW_DELHI | 0.987 | 0.996 | 1.009 |
| **TAMIL_NADU_CHENNAI** | **1.142** | **0.996** | **0.984** |
| TAMIL_NADU_COIMBATORE | 0.968 | 1.147 | 0.995 |
| **MAHARASHTRA_MUMBAI** | **1.197** | **0.952** | **0.956** |
| **GUJARAT_AHMEDABAD** | **1.161** | **1.062** | **0.945** |
| **GUJARAT_SURAT** | **1.237** | **1.087** | **0.971** |

The derived morphological features were validated through visual inspection using satellite and Google Street View imagery for Mumbai and Bengaluru. Neighborhoods in Red regions are characterized by narrower streets and potentially poor-quality housing compared to Green regions, which often feature wide streets and high-rise structures

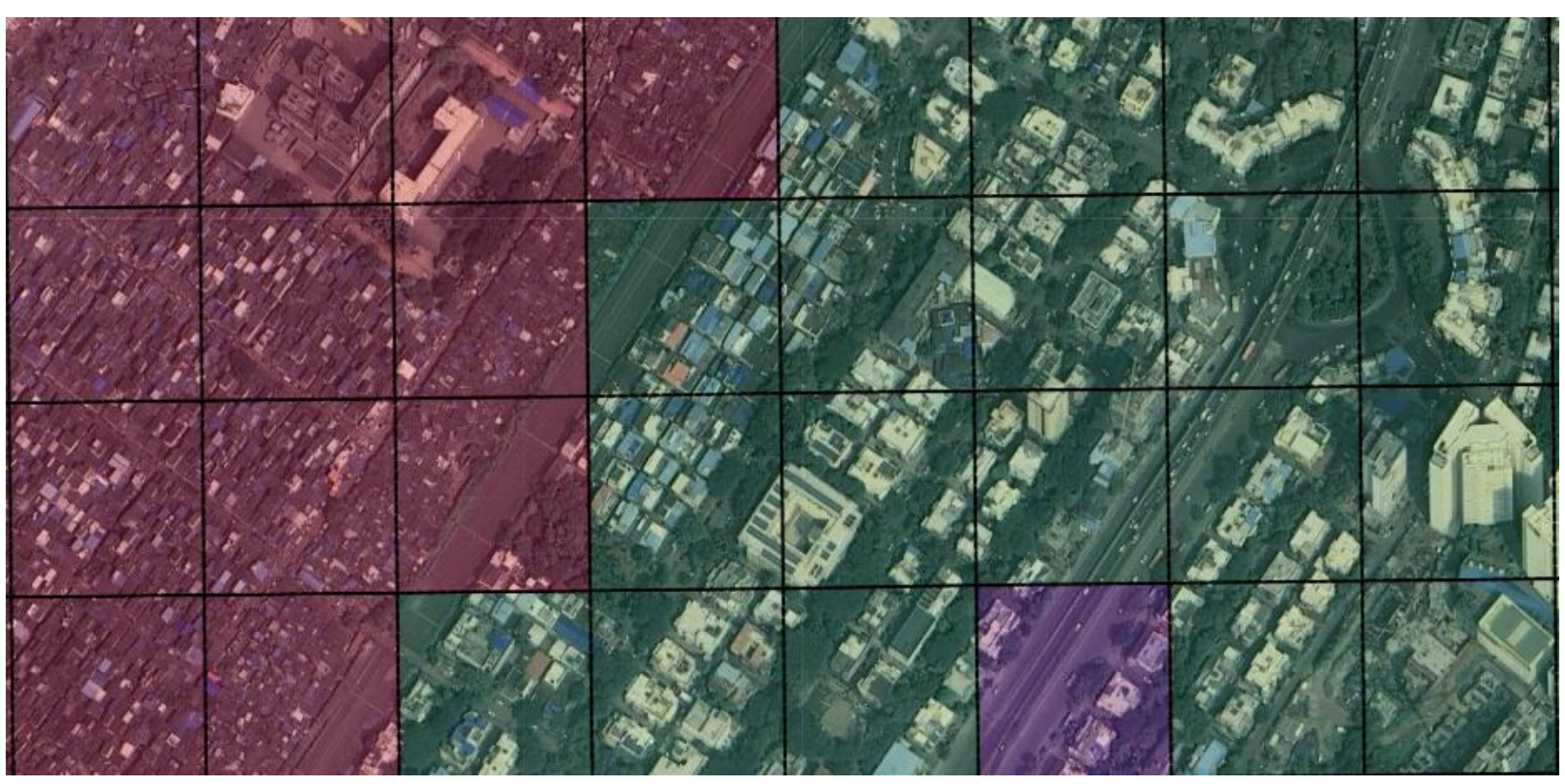

**Fig. 17.** Combined satellite views near Sion-Dharavi area, purple grids predominantly contain road infrastructure with few buildings. This visual contrast reinforces the ability to delineate neighborhoods based on morphological attributes

1 The results of all the 59 cities are mentioned in Appendix.

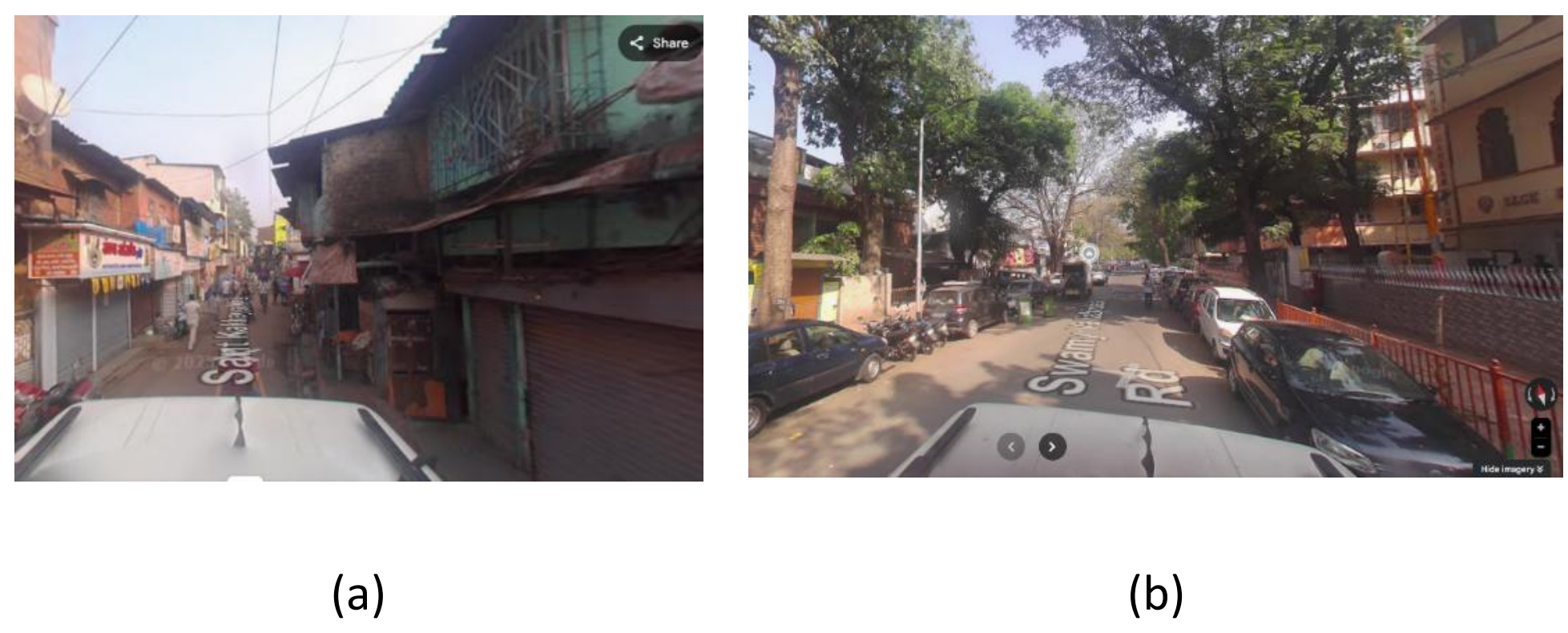

**Fig. 18.** (a) is neighborhood from Red Grid (Dharavi) and Figure (b) is from Green (Sion)

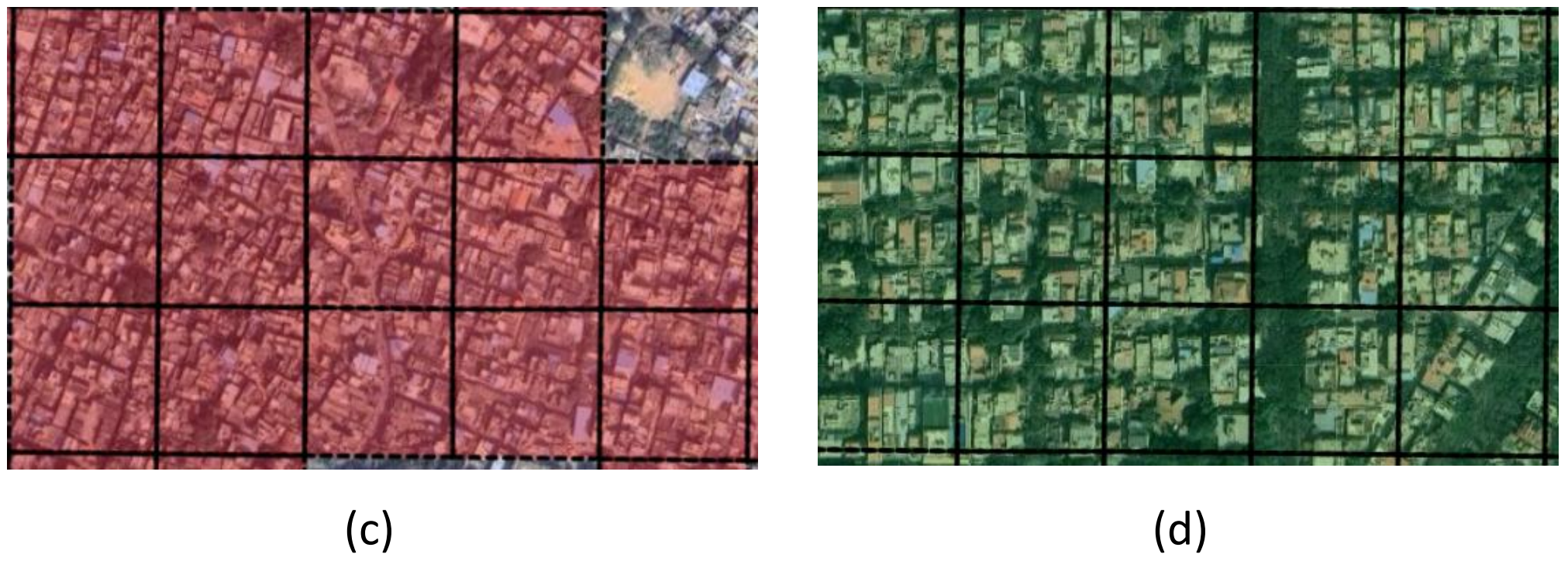

**Fig. 19.** (c) DJ Halli (d) HB Samaja Road Both images have same tiles but left one has more houses built very closely compared to the right image.

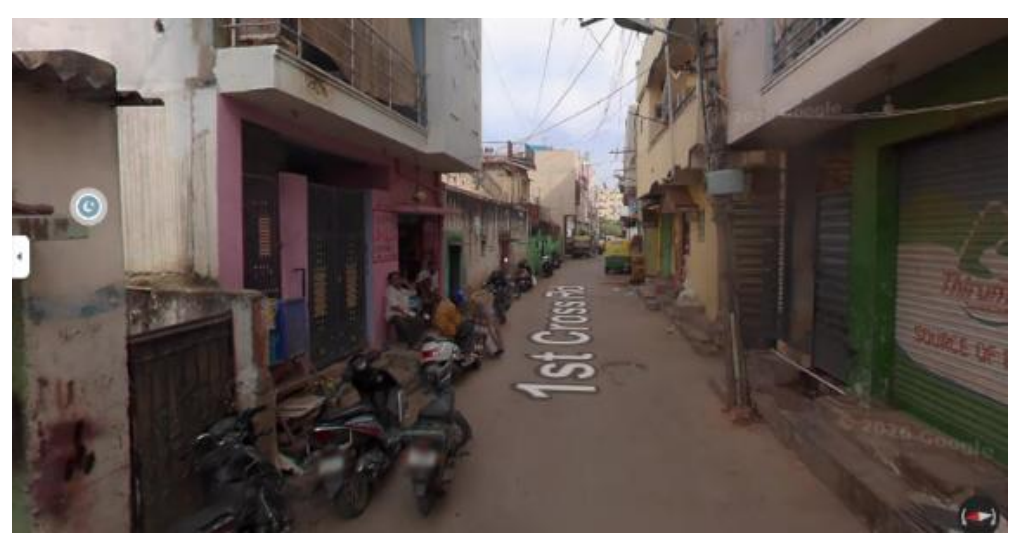

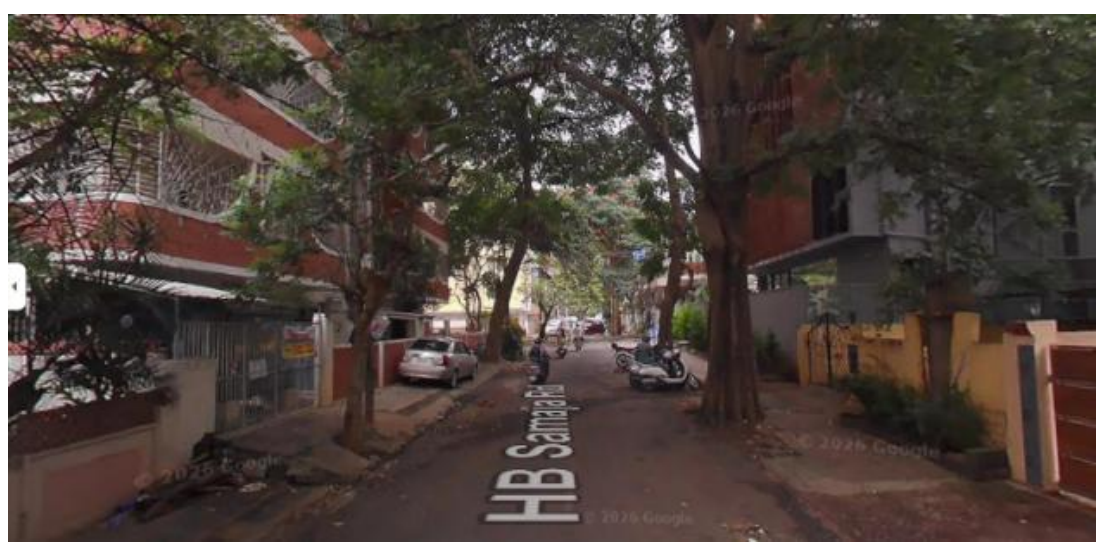

**Fig. 20.** Corresponding neighbourhood images from Fig. 19. (c) and (d).

## 5.2 Grid Validation

Affluence class validation was done using Google Street View where the streets of the grids were visited virtually. Note that the Red Grids are densely populated with narrow streets relative to the Green Grids. More validation images are provided in the Appendix section. We observe that our linear model is transferable to different cities and towns of the country.

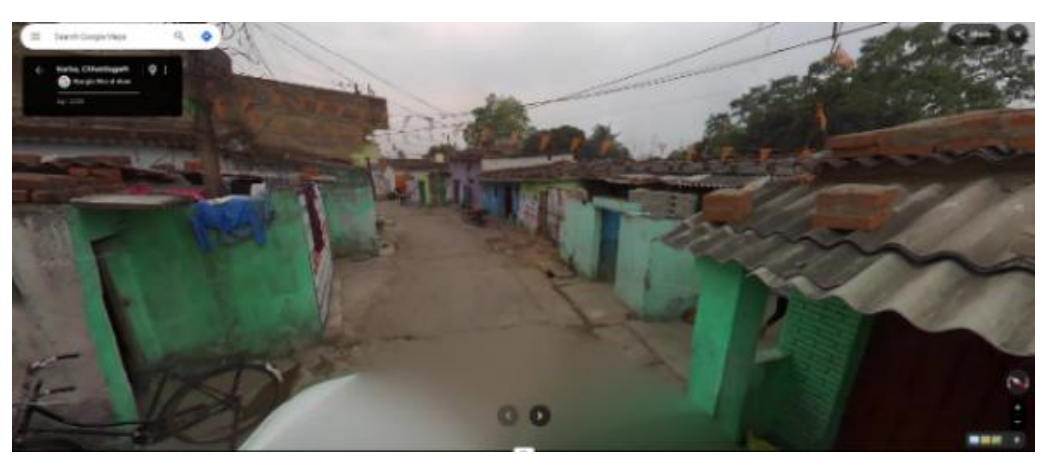

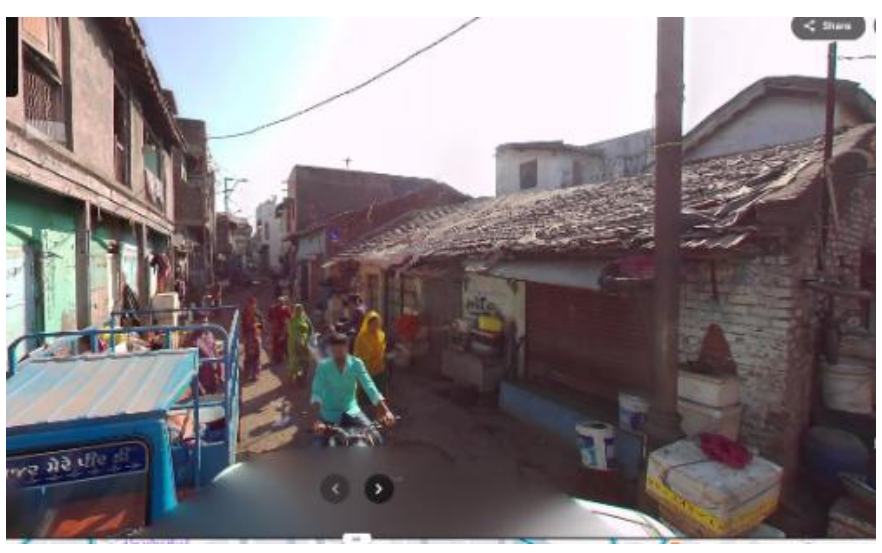

Korba Bhavnagar

**Fig. 21.** Google Street View images of Red Grids

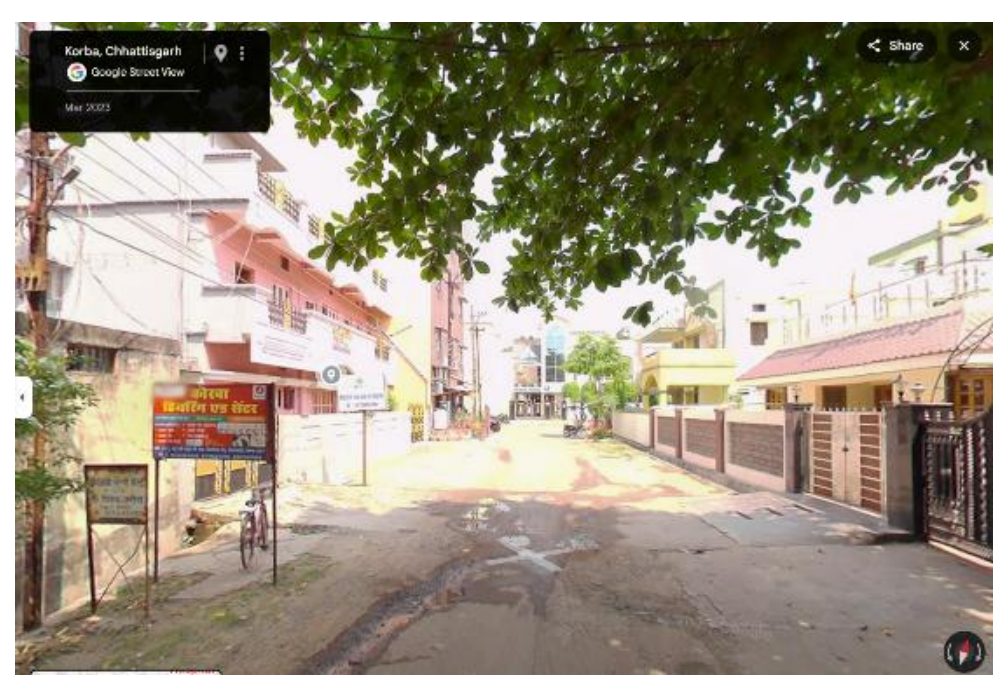

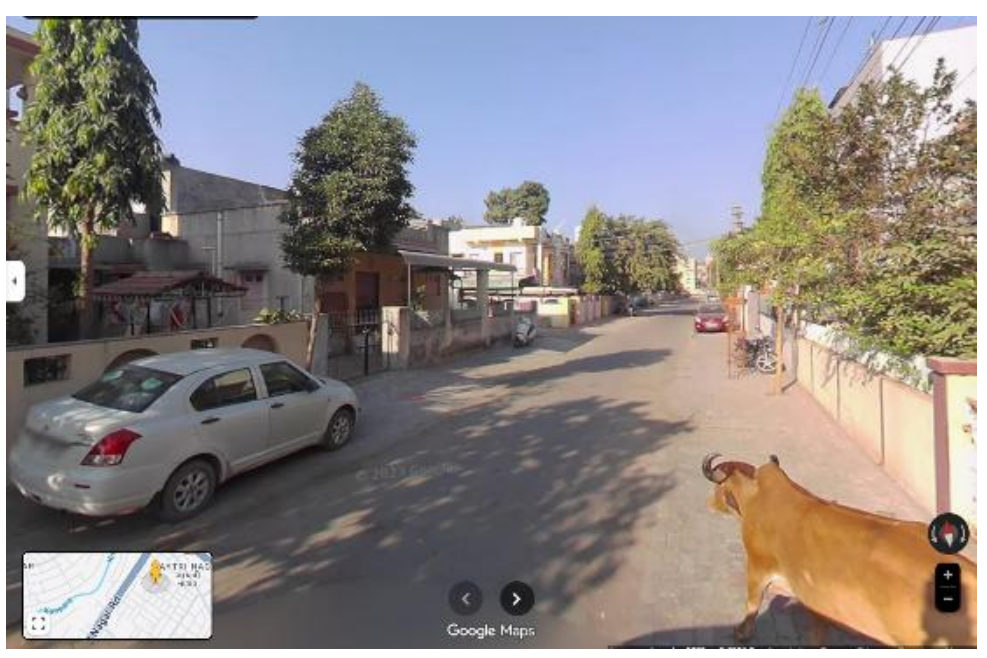

Korba Bhavnagar

**Fig. 22.** Google Street View images of Green Grids

## 5.3 Limitations

A significant challenge in scaling this methodology to a pan-India level involves the integration of rural areas using the Open Buildings dataset alone. Unlike the high-density heterogeneity of urban centers, village settlements are characterized by well-spaced and uniformly distributed housing. We used the sub-district administrative boundary by Survey of India [24] of Korba city to create grids across the region where the boundary includes the nearby villages falling under the Korba administration. The grid classification would be based on $5^{th}$ and $9^{th}$ percentile values of entire grids combined as explained previously. The Korba city area had a good neighborhoods segregation into red, amber, green classes but due to lower density in the villages all of the grids have higher score which was greater than 9th percentile value hence all are marked as green masking the potential socio-economic variations.

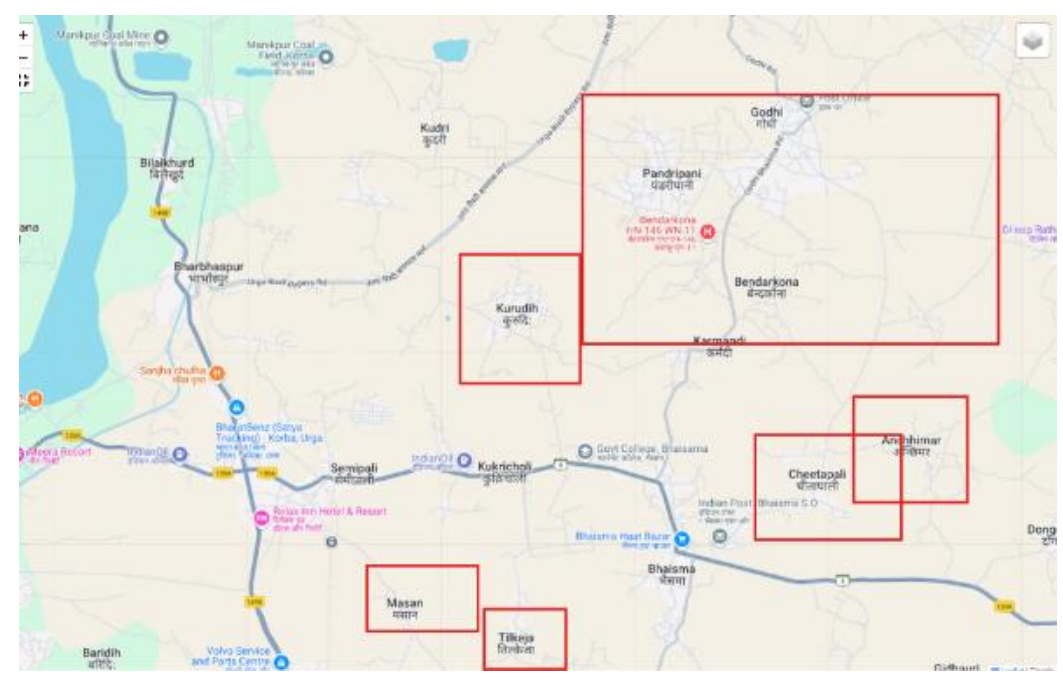

**Fig. 23.** Nearby villages of Korba City

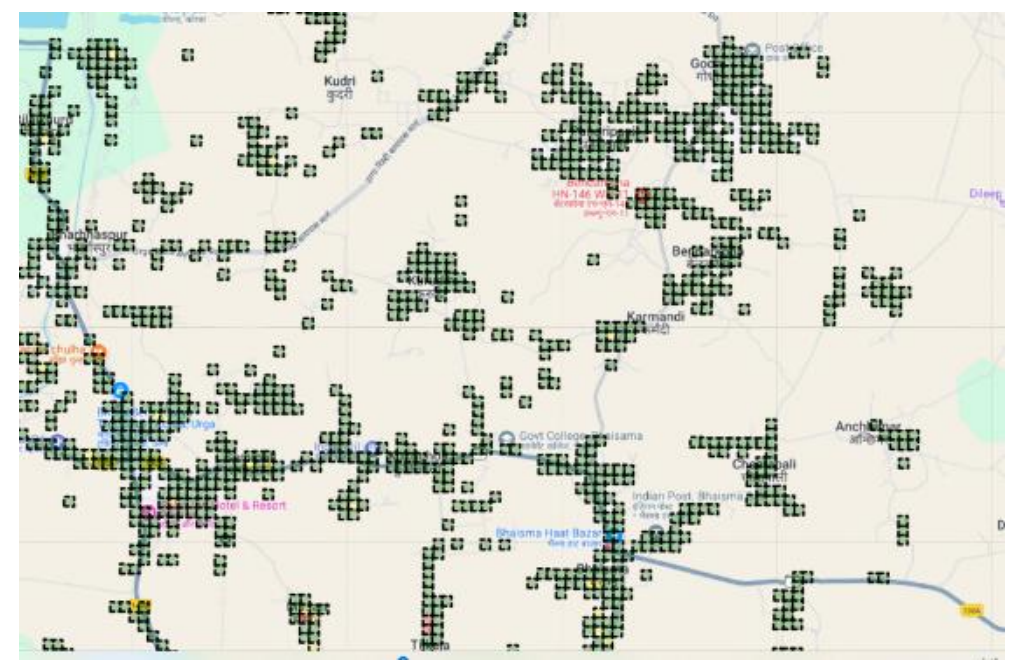

**Fig. 24.** Grids constructed

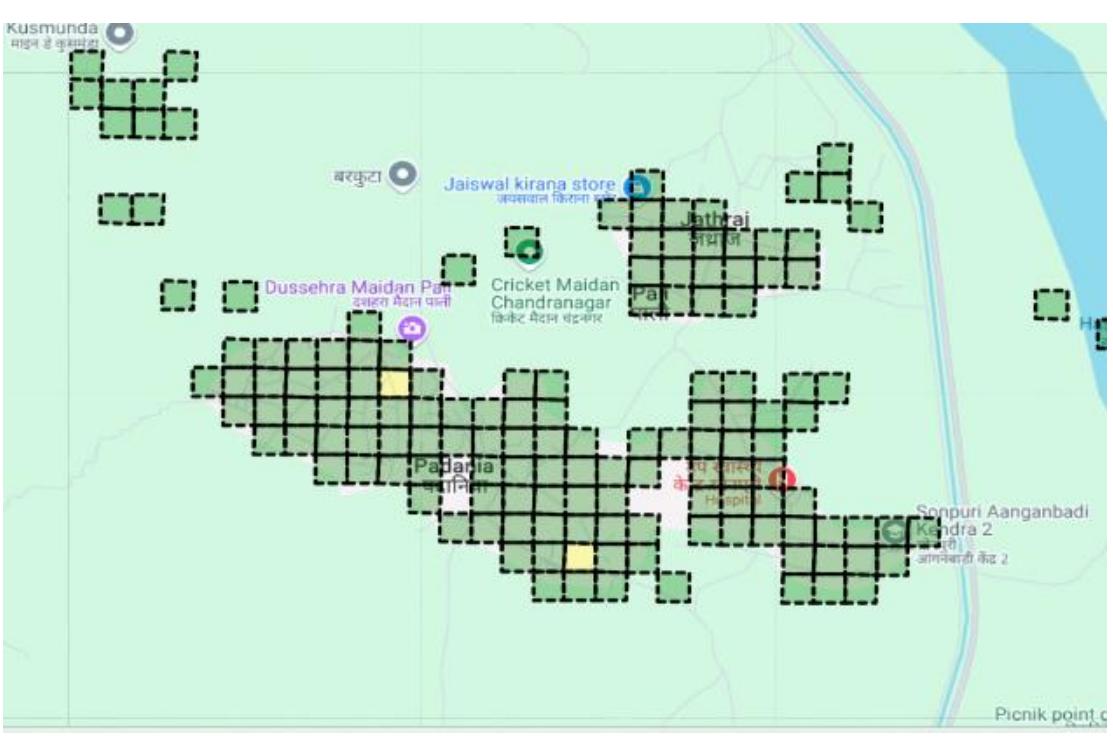

**Fig. 25.** Most of the grids are having high score

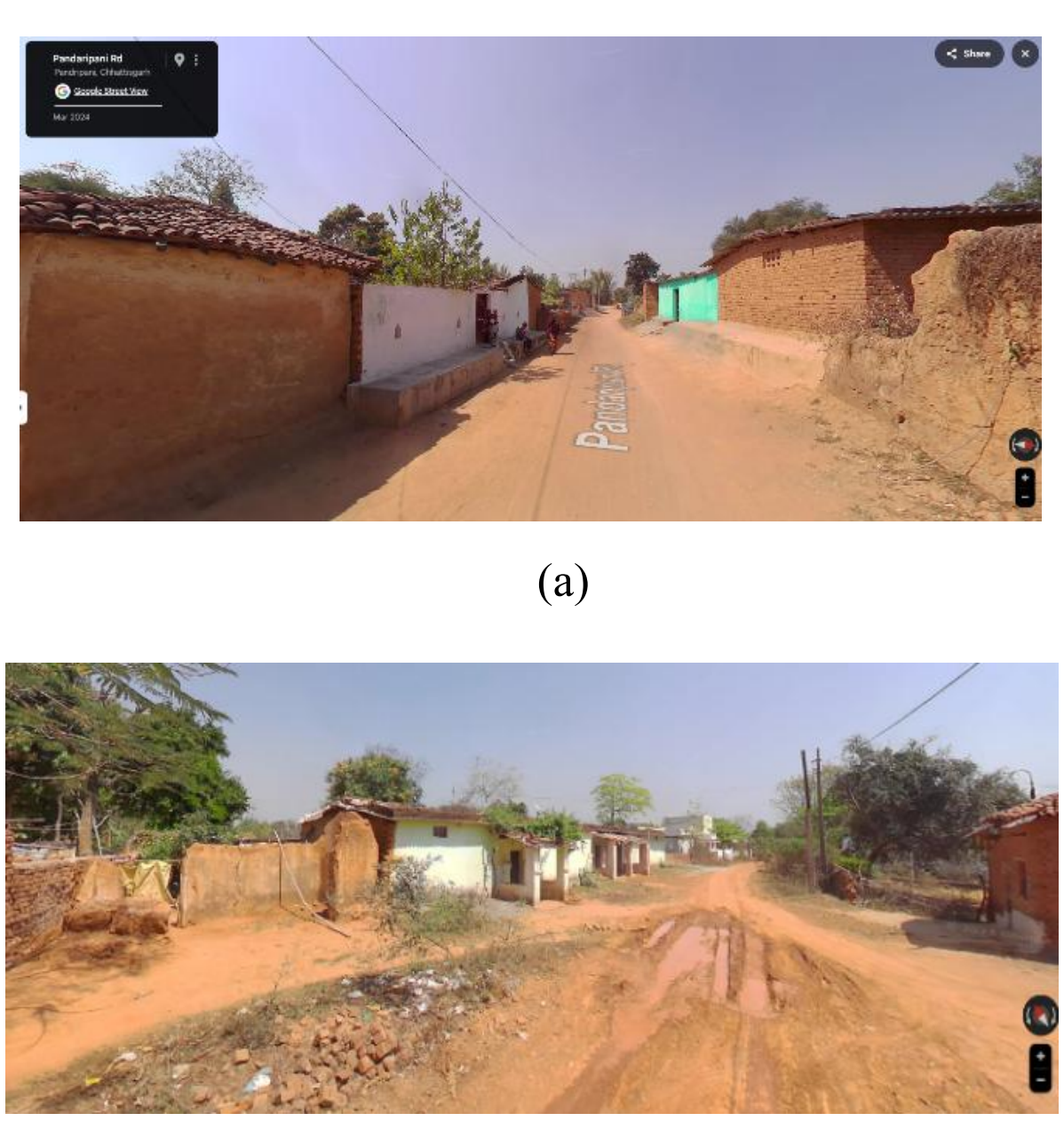

(a)

(b)

**Fig. 26.** Figure (a) is “Yellow” and (b) is “Green” grid which are very similar. Pandaripani village.

## 6 Conclusions

In this paper, we study relationship between lifestyle index of the locality, developed using morphological features derived from satellite data and the credit behavior of the residents for different cities and towns of India. Urban areas face an ever-present challenge of space constraints due to the continuous increase in migration every year. They are characterized by non-uniform spatial distribution of building density, driven by the varying demand for commercial office space and localized accessibility factors; this morphological heterogeneity serves as a robust proxy for delineating neighborhood boundaries and defining their socio-economic character. By establishing a 100m x 100m spatial grid and developing a specialized scoring methodology, we classified neighborhoods into distinct classes: Red, Amber, Green i.e. different settlements. A strong positive correlation (can be positive or negative) of delinquency and the aforementioned classes was observed across 26 cities, which may serve as an indicator of differing lifestyle levels. Furthermore, by clustering the centroids of building geometries in Mumbai, we identified clusters that significantly overlapped with known informal settlements. Extending this approach to more cities requires city boundaries which currently is a challenge. The publicly available administrative boundaries often fail to account for the latest developments at city outskirts. This may be overcome by using clustering of building centroids to define metropolitan boundaries. The current approach is limited to urban areas and failed to extend to rural areas such as villages due to uniform density distribution of the houses. Future research could focus on leveraging the nighttime light intensity data which has been found to be a proxy for economic activities in urban areas [9]. While the current Open Buildings dataset provides 2D information leveraging building height data from Open Buildings 2.5D Temporal [25] could give us 3D data and reduce misclassification. Finally, the integration of government datasets could enhance the accuracy of the identified lifestyle levels. Specifically, Income Tax data provides insights into earning capacity, GST data signals economic and consumption activity, and Property Tax data could correlate highly with the lifestyle levels associated with planned and spacious layouts.

## 7 Data Availability

The credit behavior dataset used during the current study are proprietary and were used under license from L&T Finance (LTF). Consequently, these data are not publicly available. However, data may be made available from the authors upon request and with the approval of L&T Finance.

## 8 Acknowledgements

The authors are grateful to L&T Finance for providing access to the anonymized credit behavior datasets and for supporting this research initiative. We also express our deepest thanks to Kamlesh Kumar for his continuous mentorship, guidance, and rigorous review throughout the development of this framework; his detailed feedback critically helped us and improved the final manuscript.

# 10 Appendix

## 10.1 Delinquency 59 Cities

The cities are ordered based on customer count varying from lakhs to a minimum of 10,000 customers. Note that the rows in bold are having trend.

| CITY | Red Grid | Amber Grid | Green Grid |
|---|---|---|---|
| **KARNATAKA_BENGALURU** | 1.249 | 1.068 | 0.966 |
| DELHI_NEW_DELHI | 0.987 | 0.996 | 1.009 |
| **TAMIL_NADU_CHENNAI** | 1.142 | 0.996 | 0.984 |
| **TELANGANA_HYDERABAD** | 1.104 | 1.132 | 0.978 |
| **MAHARASHTRA_MUMBAI** | 1.197 | 0.952 | 0.956 |
| **GUJARAT_AHMEDABAD** | 1.161 | 1.062 | 0.945 |
| **GUJARAT_SURAT** | 1.237 | 1.087 | 0.971 |
| WEST_BENGAL_KOLKATA | 1.093 | 0.900 | 1.012 |
| PUNJAB_LUDHIANA | 0.993 | 1.027 | 0.995 |
| **MAHARASHTRA_PUNE** | 1.373 | 1.046 | 0.934 |
| **MADHYA_PRADESH_INDORE** | 1.064 | 0.979 | 0.976 |
| GUJARAT_VADODARA | 0.980 | 1.000 | 0.997 |
| UTTAR_PRADESH_GHAZIABAD | 1.025 | 0.908 | 1.001 |
| **BIHAR_PATNA** | 1.107 | 1.156 | 0.968 |
| **MAHARASHTRA_PIMPRI_CHINCHWAD** | 1.333 | 1.144 | 0.951 |
| **MAHARASHTRA_NAGPUR** | 1.251 | 1.050 | 0.994 |
| UTTAR_PRADESH_LUCKNOW | 1.104 | 0.896 | 1.003 |
| UTTAR_PRADESH_KANPUR | 0.981 | 1.002 | 0.972 |
| **MAHARASHTRA_NASHIK** | 1.240 | 0.974 | 0.949 |
| **RAJASTHAN_JAIPUR** | 1.206 | 1.070 | 0.956 |
| **JHARKHAND_RANCHI** | 1.288 | 0.940 | 0.936 |
| ASSAM_GUWAHATI | 1.053 | 1.254 | 0.980 |
| ODISHA_BHUBANESWAR | 0.981 | 1.115 | 1.026 |
| UTTAR_PRADESH_NOIDA | 0.971 | 1.137 | 1.021 |
| TAMIL_NADU_COIMBATORE | 0.968 | 1.147 | 0.995 |
| **MAHARASHTRA_AURANGABAD** | 1.079 | 1.051 | 1.039 |
| HARYANA_GURUGRAM | 0.951 | 0.875 | 0.992 |
| HARYANA_FARIDABAD | 0.998 | 0.976 | 1.003 |
| **JHARKHAND_JAMSHEDPUR** | 1.200 | 1.354 | 1.012 |
| **ANDHRA_PRADESH_VISAKHAPATNAM** | 1.013 | 1.091 | 0.983 |

| CITY | Red Grid | Amber Grid | Green Grid |
|---|---|---|---|
| **GUJARAT_RAJKOT** | 1.178 | 1.122 | 0.997 |
| **WEST_BENGAL_SILIGURI** | 1.322 | 1.203 | 0.958 |
| UTTAR_PRADESH_VARANASI | 0.938 | 0.938 | 0.952 |
| UTTARAKHAND_DEHRADUN | 0.849 | 0.815 | 0.991 |
| UTTAR_PRADESH_GREATER_NOIDA | 0.974 | 1.097 | 0.978 |
| **UTTAR_PRADESH_ALIGARH** | 1.154 | 1.149 | 0.933 |
| WEST_BENGAL_DURGAPUR | 1.178 | 1.903 | 0.819 |
| GUJARAT_BHAVNAGAR | 0.782 | 1.230 | 0.967 |
| UTTAR_PRADESH_MATHURA | 0.807 | 0.961 | 0.952 |
| ODISHA_ROURKELA | 1.243 | 1.279 | 0.939 |
| ODISHA_CUTTACK | 0.758 | 0.632 | 1.004 |
| MAHARASHTRA_AHMADNAGAR | 0.997 | 1.044 | 0.968 |
| TAMIL_NADU_SALEM | 0.950 | 0.997 | 1.011 |
| **HARYANA_ROHTAK** | 0.951 | 0.941 | 0.948 |
| BIHAR_MUZAFFARPUR | 1.019 | 0.706 | 0.895 |
| WEST_BENGAL_BERHAMPORE | 0.399 | 1.333 | 0.911 |
| **GUJARAT_BHARUCH** | 1.554 | 2.083 | 1.101 |
| HARYANA_REWARI | 1.126 | 0.761 | 0.955 |
| **KARNATAKA_MYSURU** | 1.213 | 1.248 | 1.001 |
| **TAMIL_NADU_MADURAI** | 1.182 | 0.990 | 0.963 |
| **MAHARASHTRA_BHIWANDI** | 1.352 | 1.365 | 1.015 |
| **UTTAR_PRADESH_AGRA** | 0.901 | 0.931 | 0.891 |
| TAMIL_NADU_TIRUPPUR | 0.946 | 1.108 | 0.968 |
| **ANDHRA_PRADESH_VIJAYAWADA** | 1.464 | 1.090 | 0.945 |
| MAHARASHTRA_SOLAPUR | 1.013 | 0.839 | 1.048 |
| **UTTARAKHAND_HARIDWAR** | 1.269 | 1.059 | 0.932 |
| **ANDHRA_PRADESH_GUNTUR** | 1.008 | 0.945 | 0.972 |
| BIHAR_MOTIHARI | 0.780 | 0.689 | 0.962 |
| GUJARAT_ANAND | 0.987 | 0.765 | 0.980 |

## 10.2 Google Street View Validation

The below images depict different neighborhoods captured during the virtual street tour of the city.

### 8.2.1 Surat

#### *8.2.1.1 Red Grids*

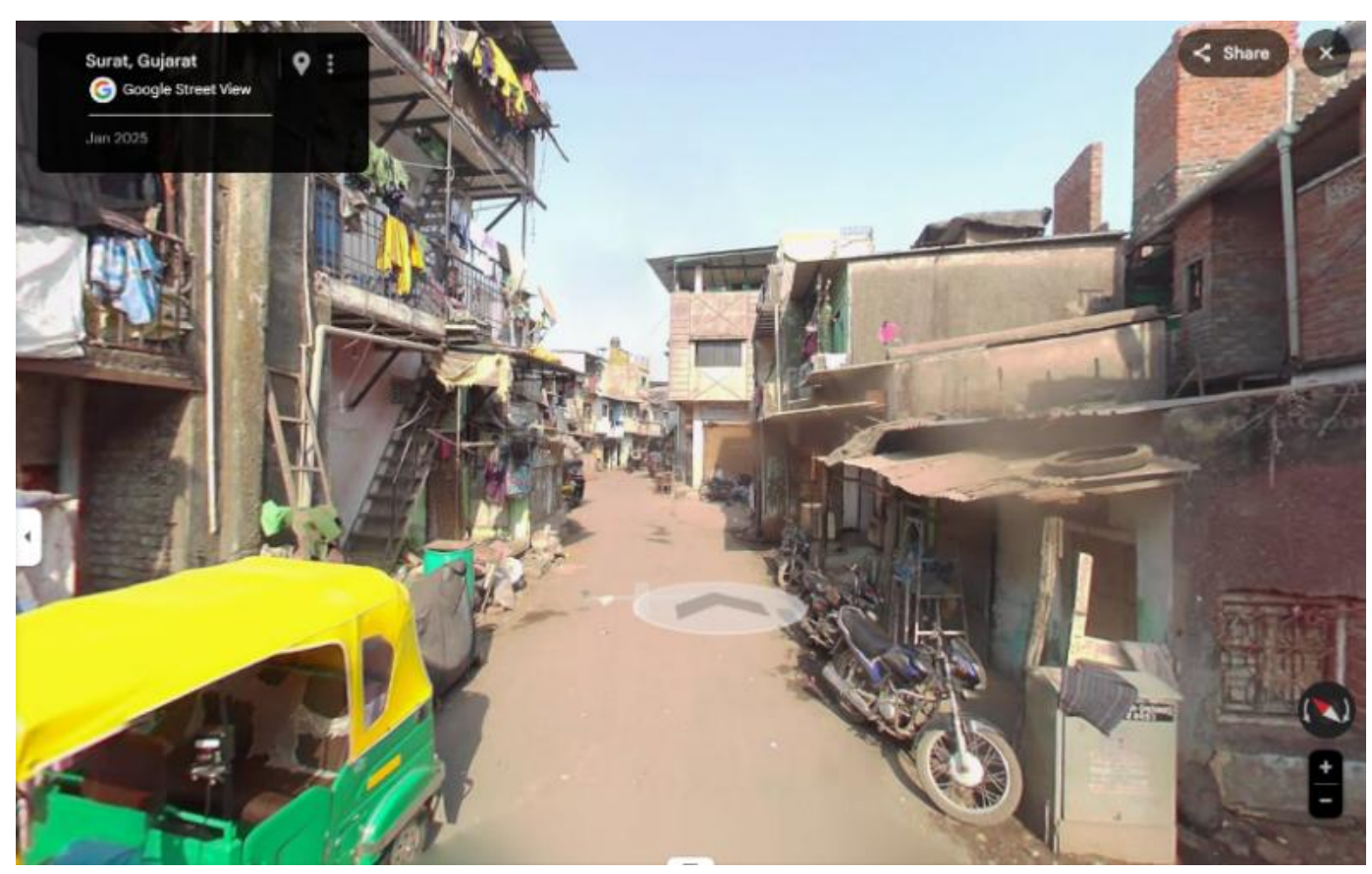

**Fig. 27.** Red Grid: Azad Chowk, Limbayat, 395012 (21.181555, 72.853148)

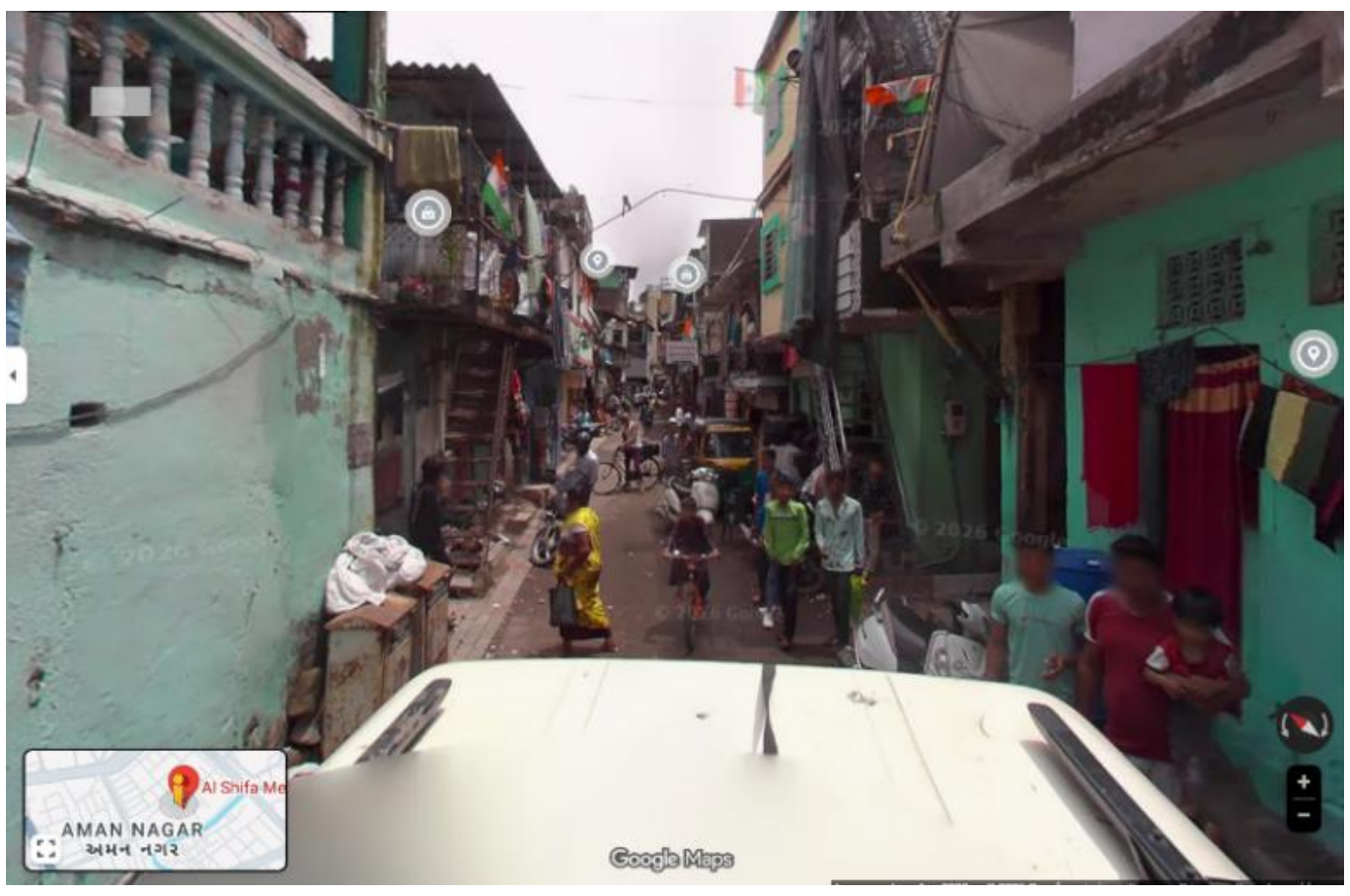

**Fig. 28.** Red Grid: Mandarwaja, near Padma Nagar, 395002 (21.182916, 72.837146)

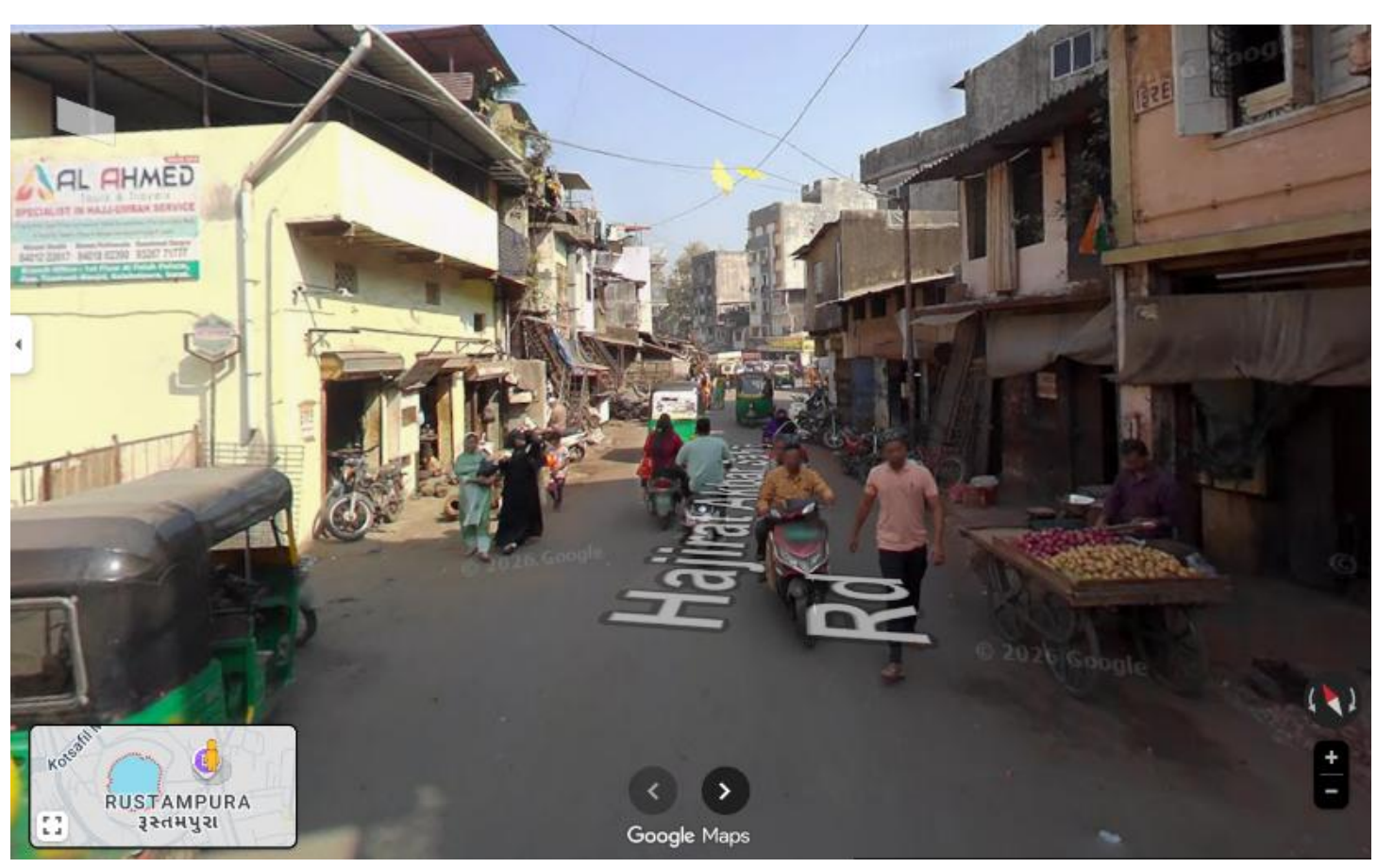

**Fig. 29.** Red Grid: Hajirat, Akbar Sahib, Rd 395002 (21.189015, 72.832081)

#### *8.2.1.2 Green Grids*

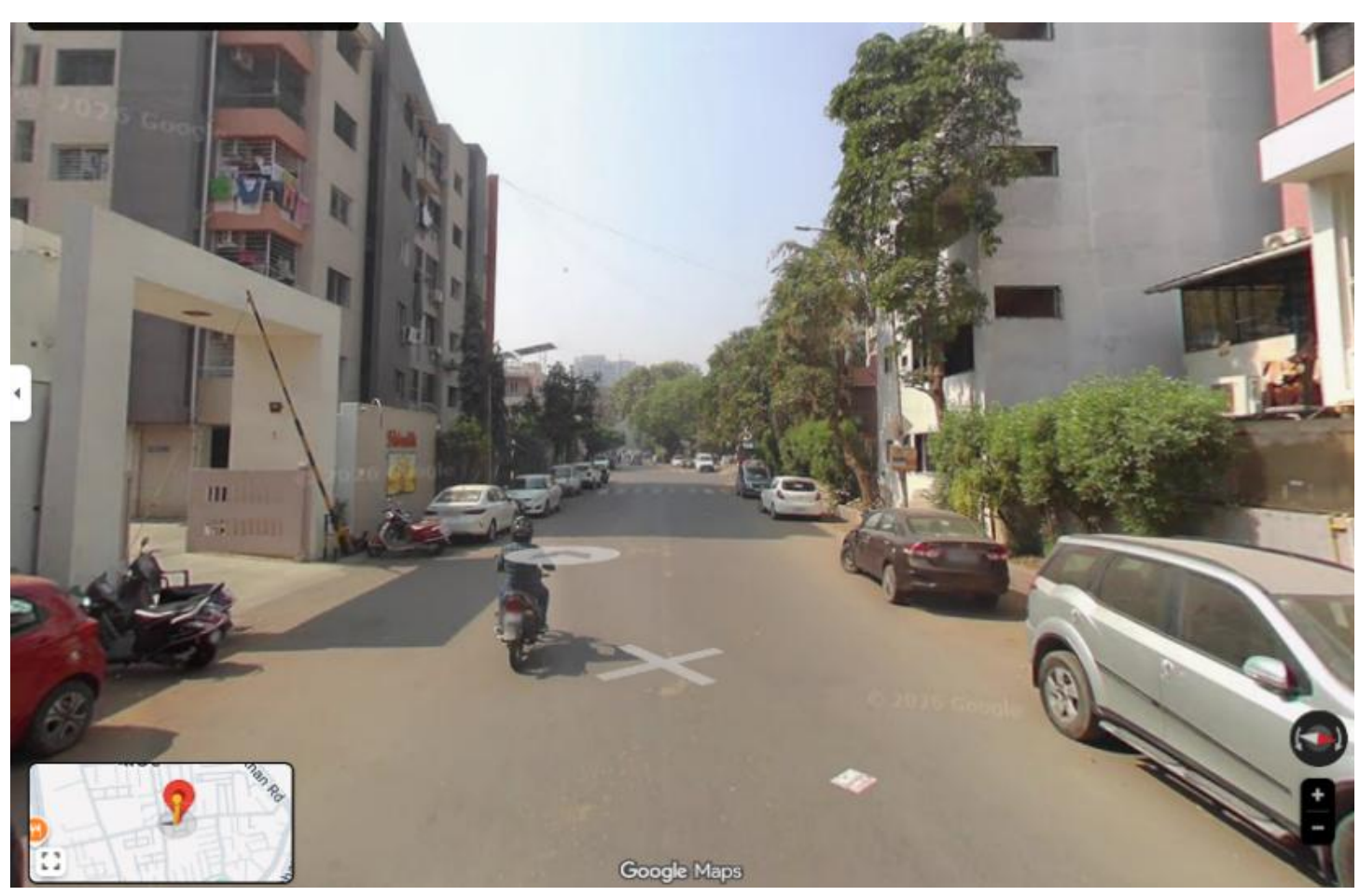

**Fig. 30.** Green Grid: New City Light, (21.150577, 72.802592)

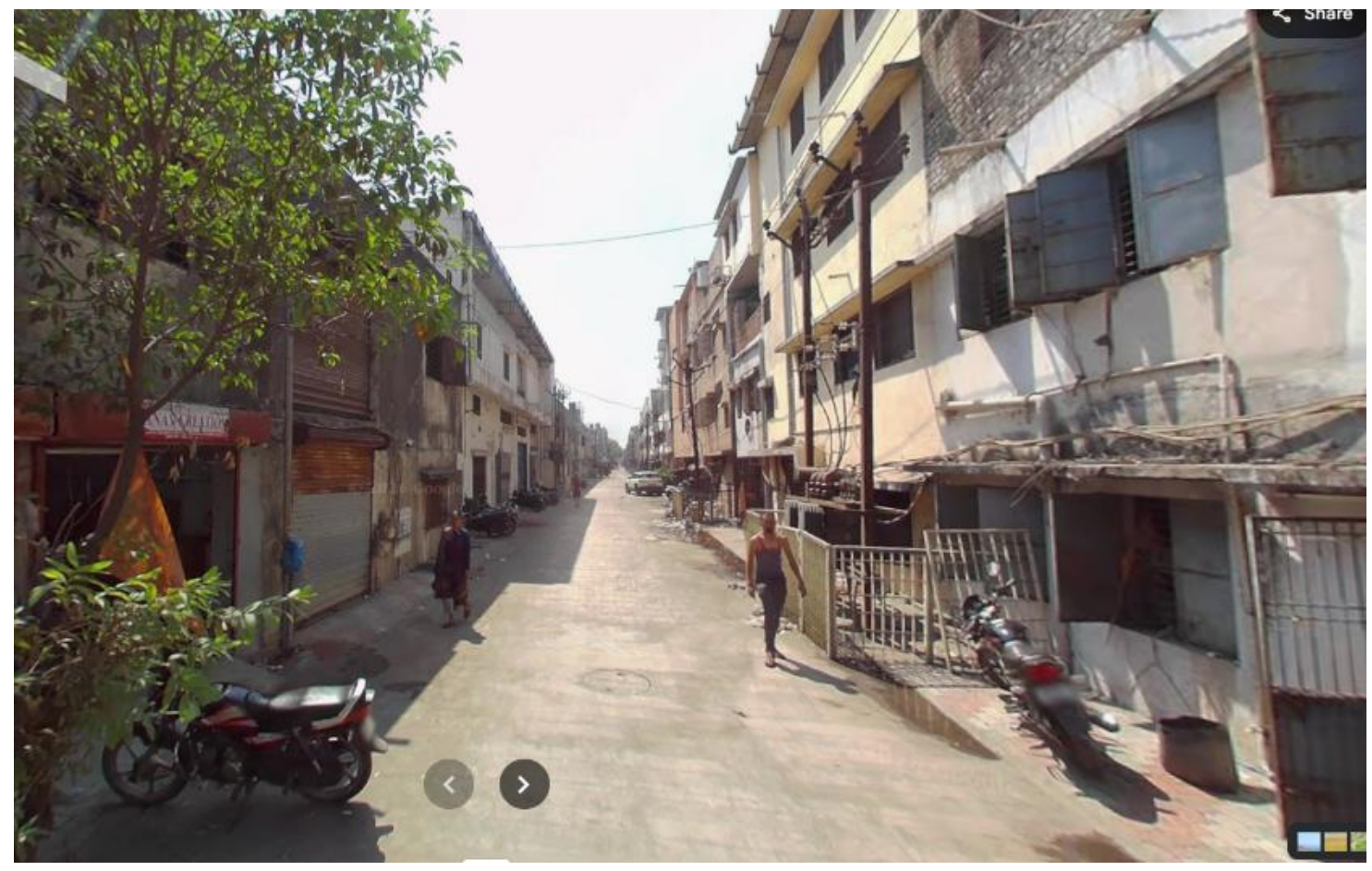

**Fig. 31.** Green Grid: Navin Flourine Colony, (21.133505, 72.858851)

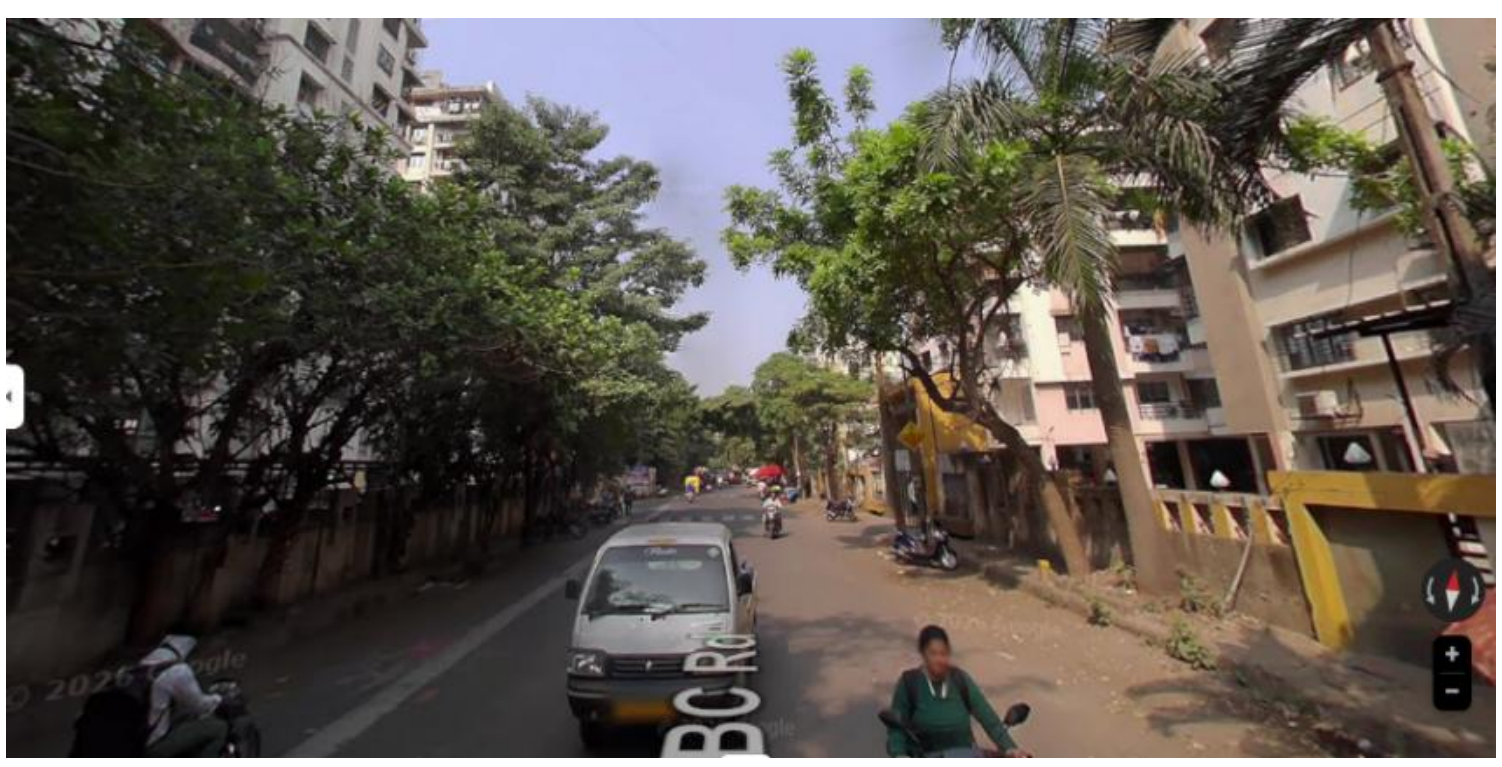

**Fig. 32.** Green Grid: Swastik Residency Road, Adajan Gam (21.188802, 72.783121)

### 8.2.2 Cuttack

*8.2.2.1 Red Grids.*

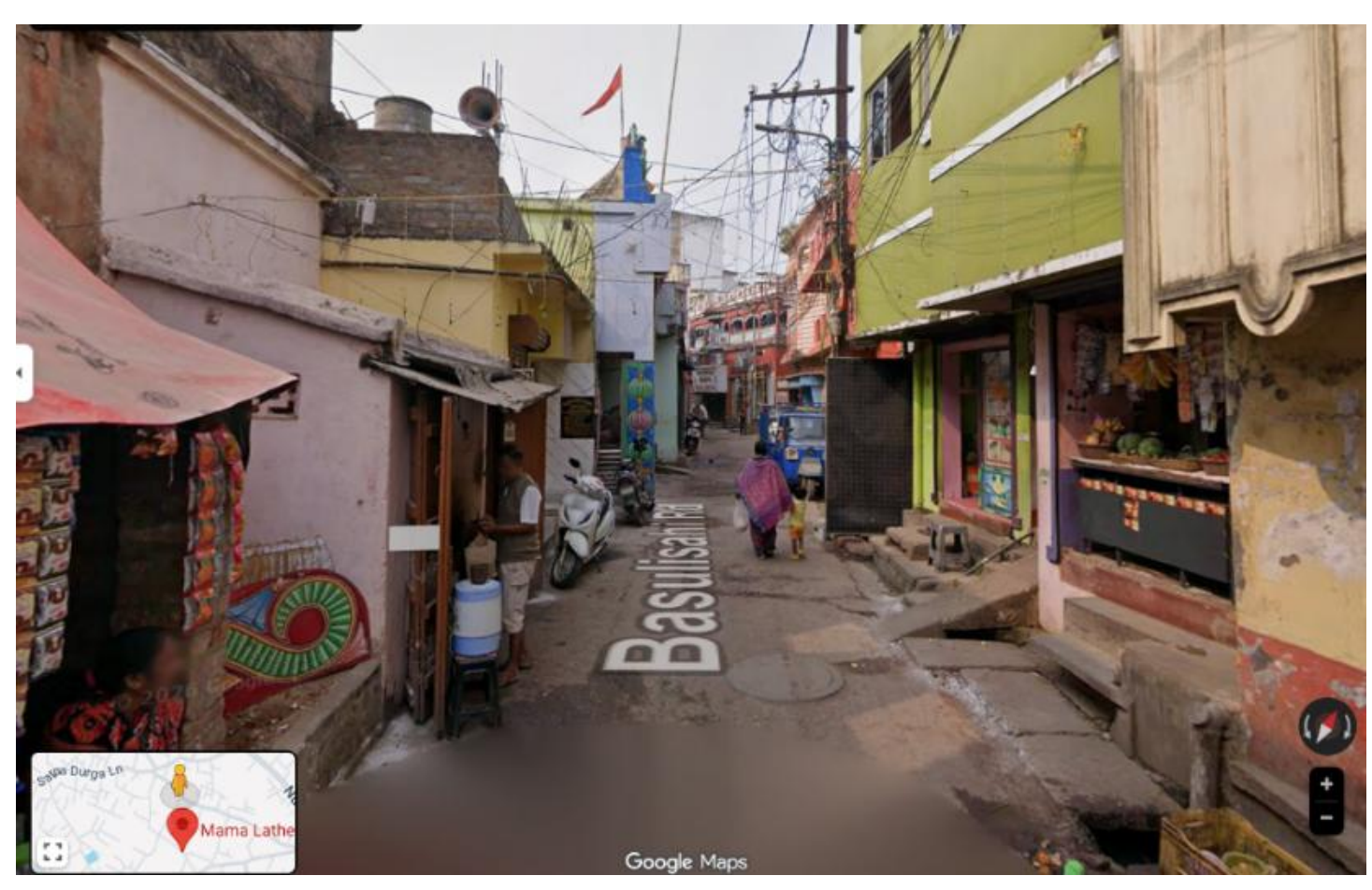

**Fig. 33.** Red Grid: Hemalata Bhawan, Basuli Sahi Road (20.474306, 85.880481)

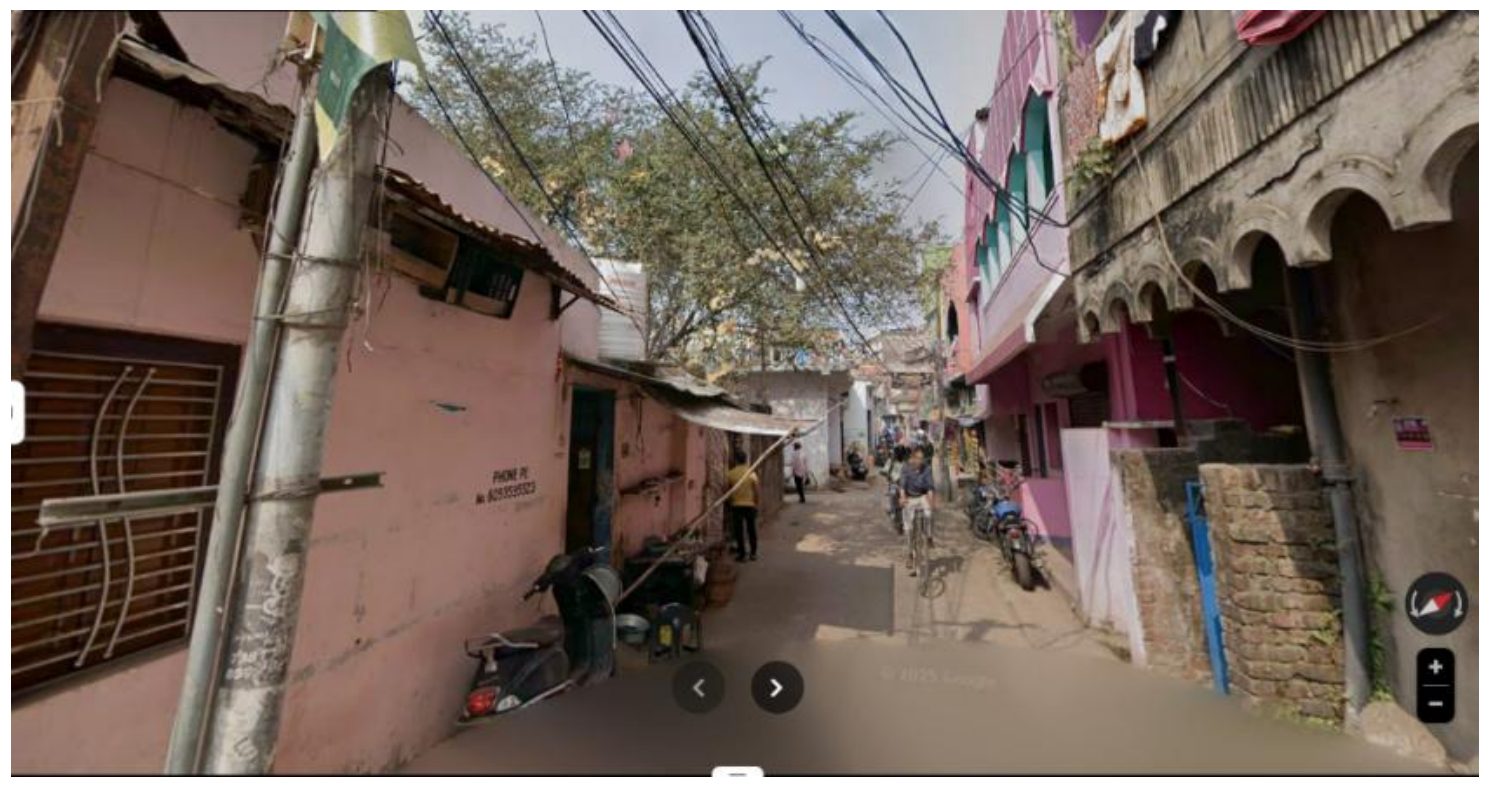

**Fig. 34.** Red Grid: Mohmaddia Bazar (20.472304, 85.855938)

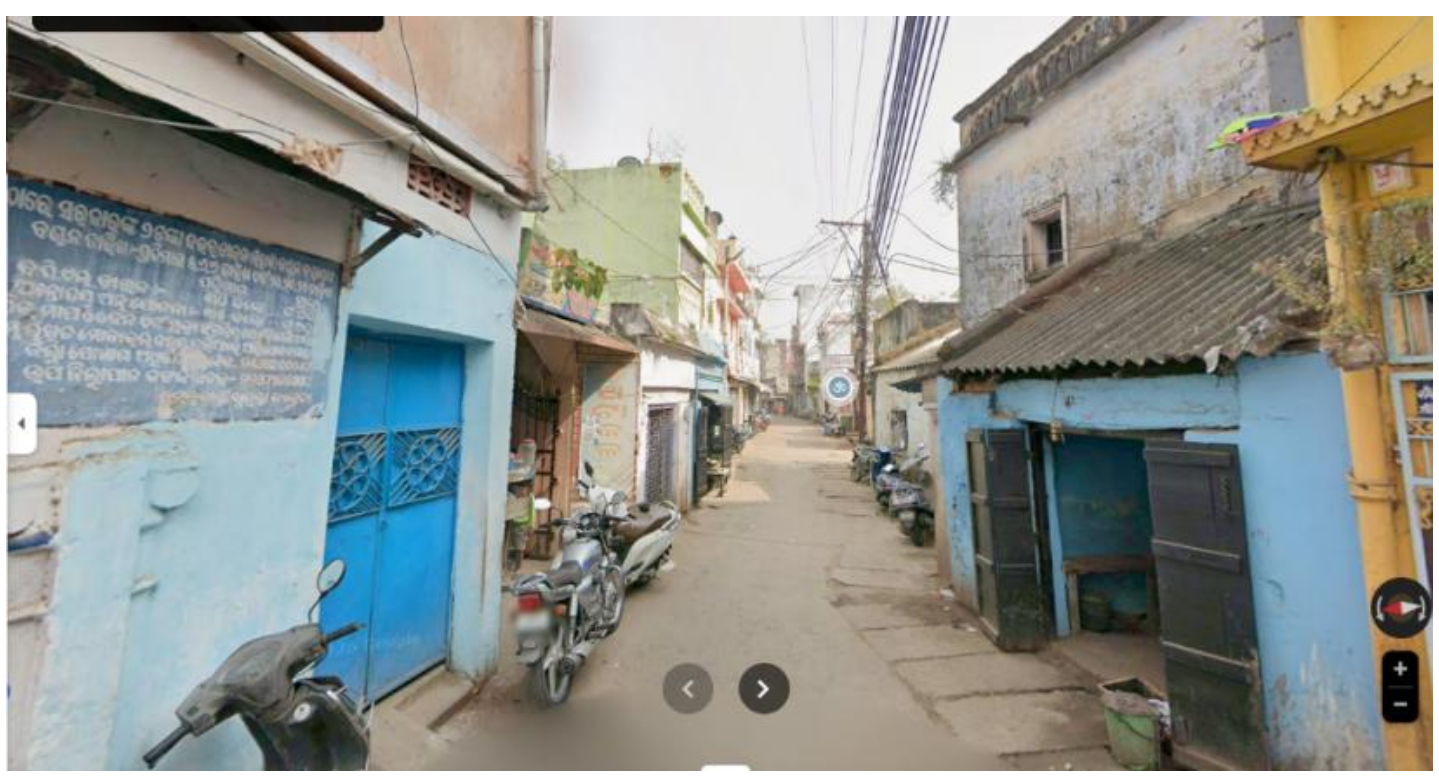

**Fig. 35.** Red Grid: Sutahat (20.471195, 85.865693)

*8.2.2.2 Green Grids*

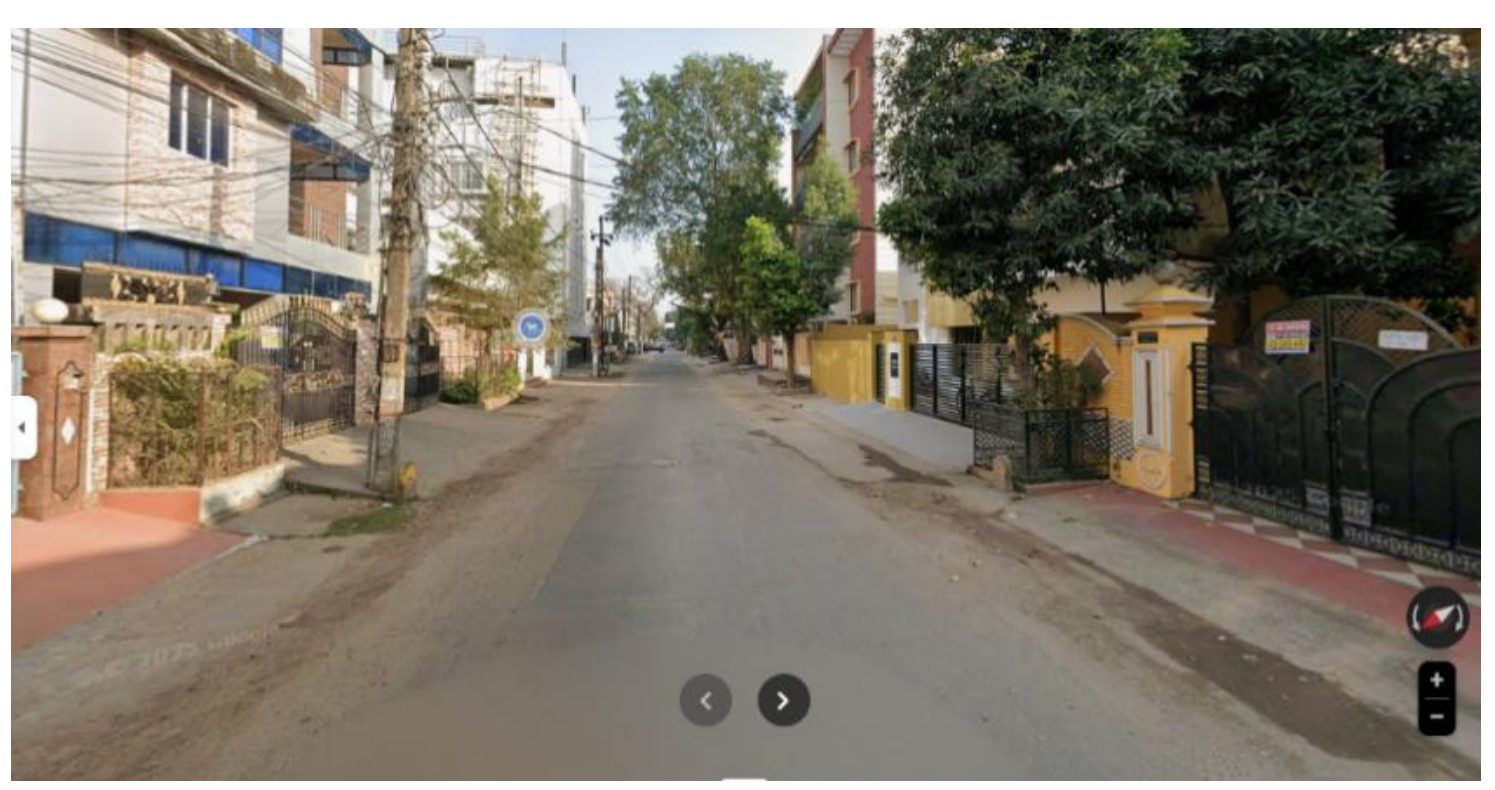

**Fig. 36.** Green Grid: CDA Sector VI (20.478190, 85.837216)

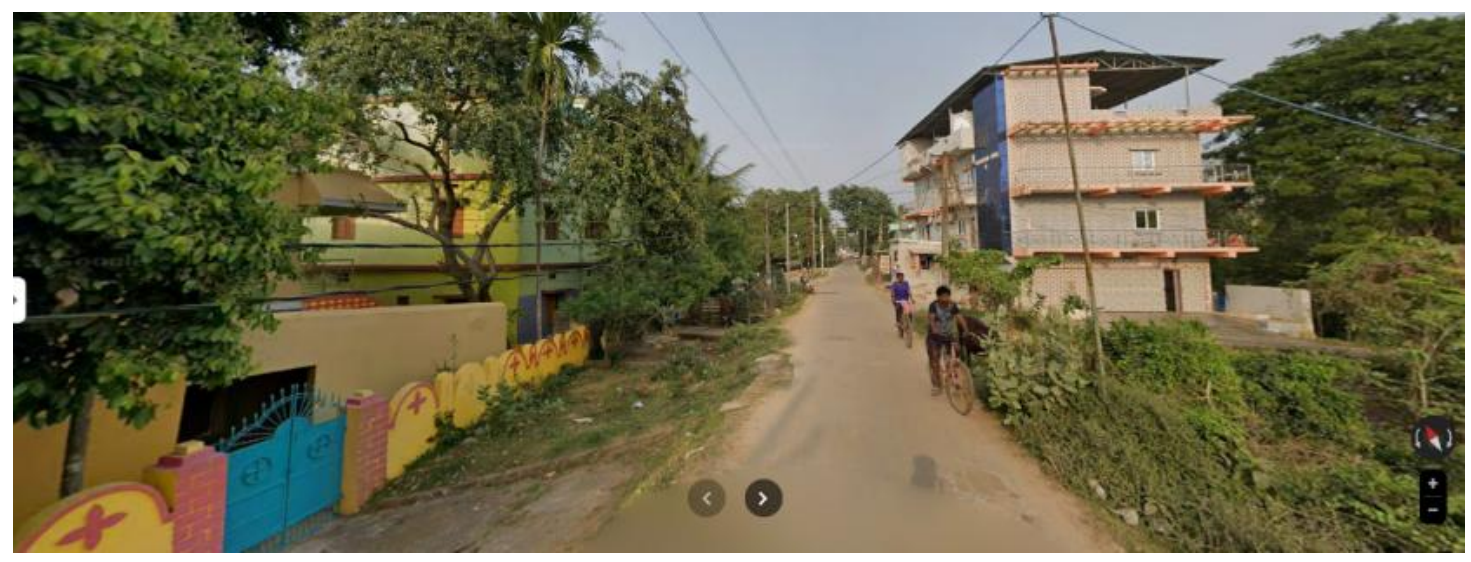

**Fig. 37.** Green Grid: Sartol (20.437774, 85.922749)

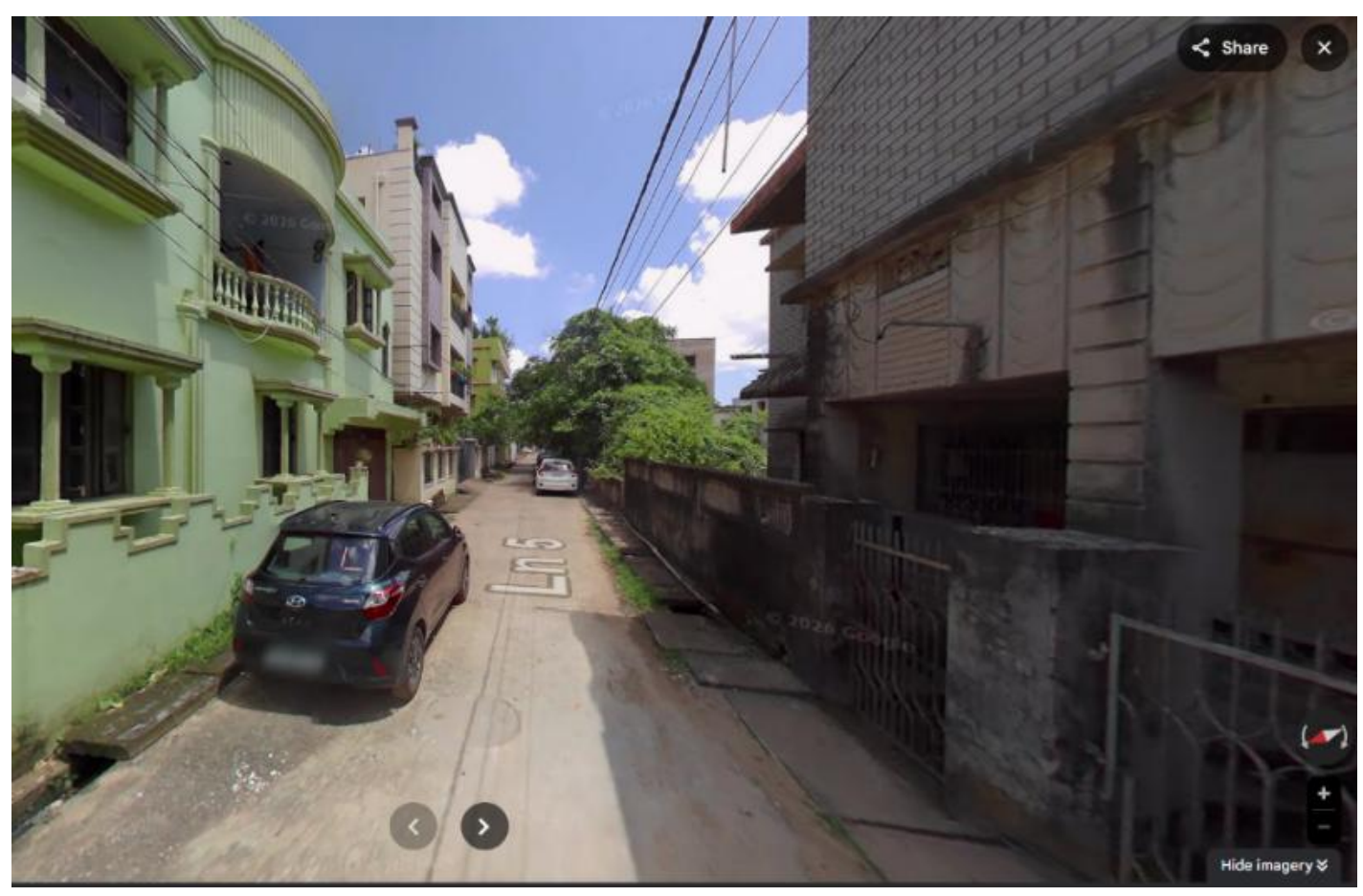

**Fig. 38.** Green Grid: Mahanadi Vihar (20.467107, 85.911049)